\documentclass[lettersize,journal]{IEEEtran}

\usepackage{amsmath,amssymb,amsfonts,pifont}
\usepackage{hyperref,graphicx}
\usepackage{tabu,float,textcomp}
\usepackage{multirow,dcolumn,hhline}
\usepackage{color,booktabs,relsize}
\usepackage{adjustbox}
\usepackage{gensymb}
\usepackage{mathtools,cuted}
\usepackage{multicol}
\usepackage{cite}
\usepackage{algorithmic}
\usepackage{booktabs}
\usepackage{bm}
\usepackage[subrefformat=parens,labelformat=parens,caption=false]{subfig}
\usepackage{makecell}
\usepackage{pifont}
\newcommand{\cmark}{\ding{51}} 
\newcommand{\xmark}{\ding{55}} 

\def\BibTeX{{\rm B\kern-.05em{\sc i\kern-.025em b}\kern-.08em
    T\kern-.1667em\lower.7ex\hbox{E}\kern-.125emX}}
\begin{document}
\title{Dual-Domain Fusion for Semi-Supervised Learning}

\author{Tuomas Jalonen,
Mohammad Al-Sa'd, \IEEEmembership{Senior Member, IEEE},
Serkan Kiranyaz, \IEEEmembership{Senior Member, IEEE},
and Moncef Gabbouj, \IEEEmembership{Fellow, IEEE}
\thanks{Tuomas Jalonen, Mohammad Al-Sa'd, and Moncef Gabbouj are with the Faculty of Information Technology and Communication Sciences, Tampere University, 33720 Tampere, Finland (e-mail: \href{mailto:tuomas.jalonen@tuni.fi}{tuomas.jalonen@tuni.fi}; 
\href{mailto:mohammad.al-sad@tuni.fi}{mohammad.al-sad@tuni.fi};\href{mailto:moncef.gabbouj@tuni.fi}{moncef.gabbouj@tuni.fi}).}
\thanks{Mohammad Al-Sa'd is also with the Faculty of Medicine and the Department of Clinical Neurophysiology at the University of Helsinki and Helsinki University Hospital, 00014, Helsinki, Finland (e-mail: \href{mailto:mohammad.al-sad@helsinki.fi}{mohammad.al-sad@helsinki.fi}; \href{mailto:ext-mohammad.al-sad@hus.fi}{ext-mohammad.al-sad@hus.fi}).}
\thanks{Serkan Kiranyaz is with the Department of Electrical Engineering, Qatar University, 2713 Doha, Qatar (e-mail: \href{mailto:mkiranyaz@qu.edu.qa}{mkiranyaz@qu.edu.qa}).}
}
\markboth{}
{\textit{Jalonen}\MakeLowercase{\textit{ et al.}}: Dual-Domain Fusion for Semi-Supervised Learning}
\maketitle
\begin{abstract} 
Labeled time-series data is often expensive and difficult to obtain, making it challenging to train accurate machine learning models for real-world applications such as anomaly detection or fault diagnosis. The scarcity of labeled samples limits model generalization and leaves valuable unlabeled data underutilized. We propose Dual-Domain Fusion (DDF), a new model-agnostic semi-supervised learning (SSL) framework applicable to any time-series signal. DDF performs dual-domain training by combining the one-dimensional time-domain signals with their two-dimensional time–frequency representations and fusing them to maximize learning performance. Its tri-model architecture consists of time-domain, time–frequency, and fusion components, enabling the model to exploit complementary information across domains during training. To support practical deployment, DDF maintains the same inference cost as standard time-domain models by discarding the time–frequency and fusion branches at test time. Experimental results on two public fault diagnosis datasets demonstrate substantial accuracy improvements of 8–46\% over widely used SSL methods FixMatch, MixMatch, Mean Teacher, Adversarial Training, and Self-training. These results show that DDF provides an effective and generalizable strategy for semi-supervised time-series classification.
\end{abstract}
\section{Introduction} \label{sec:introduction}

\IEEEPARstart{M}{achine} learning systems often require large amounts of labeled data for supervised learning, where all samples must be reliably annotated. However, annotation is particularly challenging in many real-world environments, especially in detecting and diagnosing faults or other anomalies, as it can be labor-intensive, dangerous or even infeasible when faults are rare or difficult to replicate \cite{chen2023deep}. To mitigate this issue, researchers have explored alternatives such as semi-supervised learning (SSL), unsupervised learning, and domain adaptation techniques \cite{qi2020small, gao2024multi}. 

Most existing time-series SSL methods treat input data as a single homogeneous view, often relying solely on the one-dimensional (1D) time-domain representation \cite{wang2024self, tang2024fault, jalonen2023bearing_fault, wang2019understanding, wang2025semi}. In some methods, the 1D time-domain signals are transformed into the 1D frequency-domain \cite{luo2024fft} or the two-dimensional (2D) time-frequency (TF) domain \cite{shi2021novel,alsad2024quadratic, qu2023adaptive, deng2021double, verstraete2020deep}. TF-based models often outperform time-domain models, especially in noisy conditions \cite{alsad2024quadratic, jalonen2023bearing_fault}. However, TF solutions are computationally intensive, as they require complex transformations and larger model architectures making them less ideal for deployment in latency-sensitive or resource-constrained environments \cite{alsad2024quadratic, jalonen2023bearing_fault, 9456035, TFBook}. This naturally raises the question: can we combine the expressive power of TF models with the efficiency of time-domain models, in a way that improves accuracy without increasing inference complexity?

In this work, we propose Dual-Domain Fusion (DDF), a novel SSL framework that jointly trains separate models operating on time-domain and time-frequency representations of time-series signals. Then, a decision fusion model is trained to combine the predictions of both classifiers. This fusion model learns to weight class logits optimally, enabling the system to generate high-quality pseudo-labels that are used to iteratively retrain the two primary models. This way, the TF model's best part, better accuracy, can be transferred to the time-domain model without sacrificing its lightweight, low-latency structure.

The proposed DDF method provides a general semi-supervised learning framework that is model-agnostic and applicable to any 1D time-series signal. It introduces a dual-domain strategy that leverages both the raw 1D time-domain signal and its derived 2D time–frequency representation to improve learning performance without any added inference complexity. To the best of our knowledge, this is the first SSL approach that employs a tri-model architecture consisting of time-domain, time–frequency, and fusion components. Our key contributions are summarized as follows:
\begin{enumerate}
\item We propose Dual-Domain Fusion (DDF), a new model-agnostic SSL framework applicable to any 1D time-series signal\footnote{\label{fn:code}The implementation will be available at \url{https://github.com/author/project}}.
\item Our deployment-aware architecture introduces no additional inference-time cost. All new samples are predicted solely from the 1D time-domain branch.
\item Experimental results on two public fault diagnosis datasets show substantial accuracy improvements of 8–46\% over FixMatch, MixMatch, Mean Teacher, Adversarial Training, and Self-training.
\end{enumerate}

The remainder of this paper is organized as follows. We review related works in Section \ref{sec:related_works} and explain our proposed DDF method in section \ref{sec:proposed_method}, including preprocessing, model architecture, the decision fusion mechanism, and deployment strategy. In section \ref{sec:experimental_validations} we describe the experimental validations, including datasets, baseline methods, implementation details, and results. Finally, we conclude and outline potential directions for future work in section \ref{sec:conclusions}.
\section{Related Works} \label{sec:related_works}
\subsection{Semi-Supervised Learning}
Many semi-supervised learning techniques have been proposed over the years. Self-training methods train a model with labeled samples and then assign pseudo-labels to unlabeled samples iteratively \cite{zheng2021self}. More recently introduced FixMatch \cite{sohn2020fixmatch} and Mixmatch \cite{berthelot2019mixmatch} fuse consistency regularization with pseudo-labeling. In Mean Teacher \cite{tarvainen2017mean}, a teacher model guides a student model's training using exponential moving average of model weights. Adversarial training approaches improve robustness by generating synthetic data or perturbations \cite{miyato2018virtual}. These methods perform well on general SSL benchmarks \cite{sohn2020fixmatch, berthelot2019mixmatch, tarvainen2017mean, miyato2018virtual}, but they are limited to a single data view or domain, leaving potential for improvement through more flexible approaches.
\subsection{Multi-View and Decision Fusion}
Some prior works have touched on the idea of fusing data domains. For example, feature-level co-training was used in \cite{yan2022cotraining}, but it was limited to simple 1D time and 1D frequency views, and they did not utilize richer 2D time-frequency representations. The co-training-based approach by \cite{liu2023temporal} and contrastive learning-based method of \cite{liang2025novel} are missing not only the TF view, but also decision fusion. Time, frequency and time-frequency features extracted from multiple sensors were all fed into a single model in \cite{razavi2018information}. A similar but more complicated setup was used in \cite{wei2023time} for general SSL benchmarks. While effective, both approaches rely on heavy preprocessing also during inference, as all the views, including the time-frequency representations, must be computed online. This limits applicability in real-time or edge deployment scenarios. Other efforts have explored decision-level fusion by combining classifier logits \cite{guo2020online}, using soft-voting \cite{xu2022intelligent}, mode-decomposition \cite{yang2023decision}, or employing class-wise disagreement metrics to guide pseudo-labeling \cite{fan2023conservative}. However, these methods have largely been confined to single-domain inputs, often only in the time domain.
\subsection{Knowledge Gap}
A summary of related semi-supervised methods and their characteristics is shown in Table \ref{tab:literature_comparison}, highlighting the absence of domain-level decision fusion and deployment-oriented architectures in existing work. These methods fall short in addressing the cross-domain fusion of time and time-frequency views in semi-supervised learning. They either miss the rich semantics available in TF representations, or they fail to support collaborative training across domains in a way that enables practical deployment. Even when TF is present, it is typically used as a mandatory input even during inference \cite{shi2021novel,alsad2024quadratic, qu2023adaptive, deng2021double, verstraete2020deep, razavi2018information, wei2023time} rather than in complementary fusion with the time-domain. The possibility of training a high-performing TF model offline and using it to enhance a lightweight time-domain model for inference deployment on the edge is not widely studied. This represents a clear gap for developing new semi-supervised solutions that combine the best of both domains while remaining deployable in real-world systems. 
\begin{table}[t]
\centering
\caption{Comparison of related semi-supervised methods.}
\label{tab:literature_comparison}
\begin{tabular}{lccccc}
\hline
\textbf{Paper} & \textbf{Year} & \makecell{\textbf{Time} \\ \textbf{Domain}} & \makecell{\textbf{Time-Frequency} \\ \textbf{Domain}} & \makecell{\textbf{Decision} \\ \textbf{Fusion}} & \makecell{\textbf{Edge} \\ \textbf{Ready}} \\
\hline
\cite{razavi2018information} & 2018  & \cmark & \cmark   & \xmark & \xmark \\
\cite{wei2023time} & 2023            & \cmark & \cmark   & \xmark & \xmark \\
\cite{ye2024semi} & 2024             & \cmark & \xmark   & \xmark & \xmark \\
\cite{wang2025semi} & 2025           & \cmark & \xmark   & \xmark & \cmark \\\cite{liang2025novel} & 2025         & \cmark & \cmark   & \xmark & \xmark \\
Ours & TBA                           & \cmark & \cmark & \cmark & \cmark \\
\hline
\end{tabular}
\vspace{2mm}
\footnotesize{
\cmark = supported, \xmark = not used.
}
\end{table}

\section{Proposed Method} \label{sec:proposed_method}
\begin{figure*}[!t]
\centerline{\includegraphics[width=\textwidth]{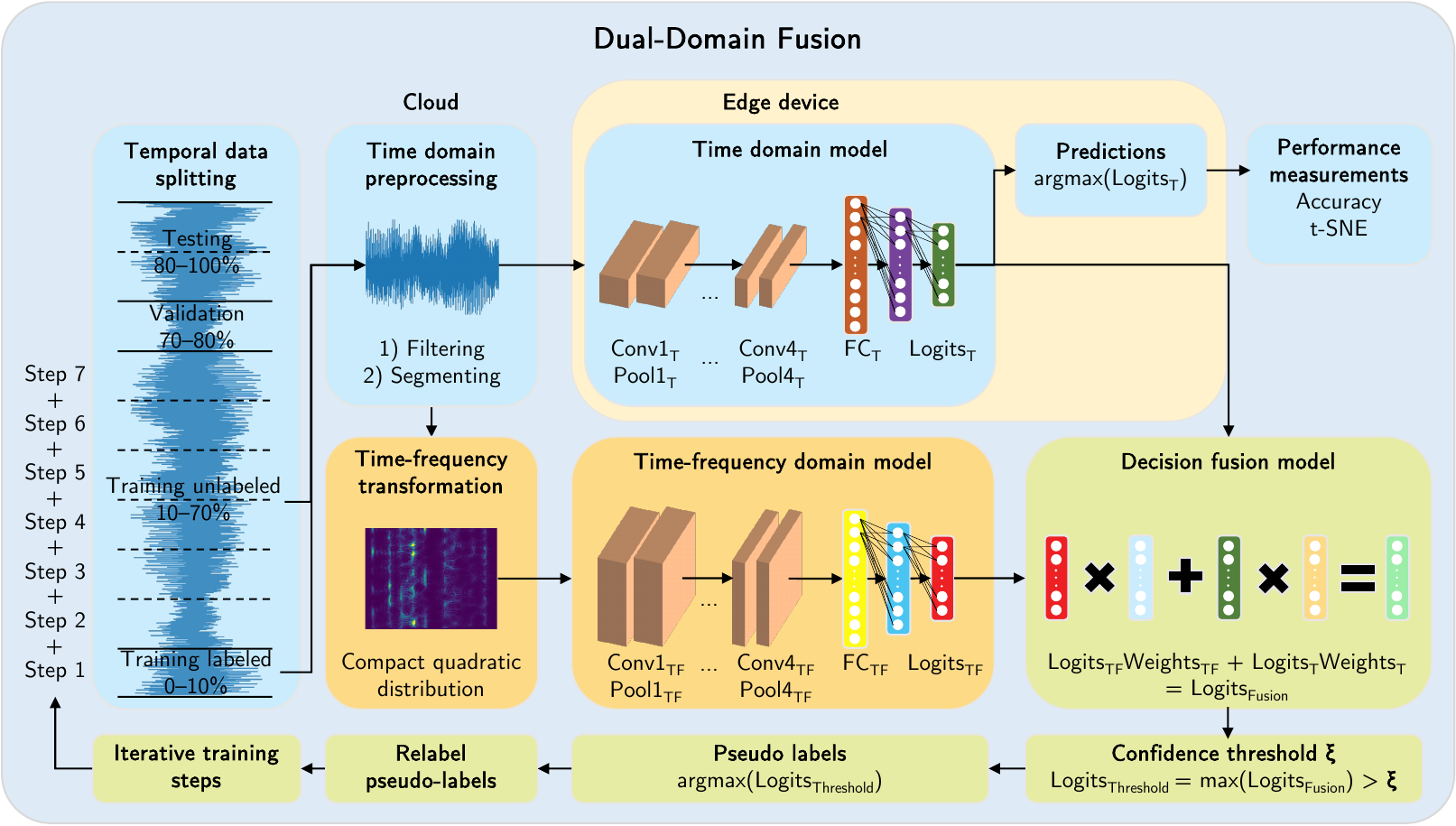}}
\caption{The overall schematic of our proposed Dual-Domain Fusion method that combines the time and time-frequency models with a decision fusion model to generate more accurate and reliable predictions.}
\label{fig:framework_overall}
\end{figure*}

The proposed Dual-Domain Fusion method is illustrated in Fig. \ref{fig:framework_overall}. Our methodology assumes a practical case where a large set of collected data is available for classification purposes; however, only a small subset of these data samples is manually labeled. In other words, few data samples can be used directly for training in a supervised fashion, while the remaining cannot be naively used and require further processing before utilization, i.e., semi-supervised learning. Besides, our technique fuses time and time-frequency (TF) classifiers to improve their overall performance, reliability, and versatility for real-time operation.
\subsection{Dual-Domain Fusion}
The Dual-Domain Fusion technique begins with preprocessing the labeled samples in the time domain and estimating their joint time-frequency representation, generating two separate data views: a 1D time domain view and a 2D time-frequency domain view. Separate classifiers are trained in a supervised manner using each of these labeled data views in Step 1 in Fig. \ref{fig:framework_overall}. Additionally, a decision fusion model is trained using the outputs of these classifiers to enhance accuracy and robustness. Subsequently, the trained models are applied to the next subset of the remaining unlabeled samples to generate pseudo-labeled predictions. These pseudo-labeled samples are incorporated into the original training set if the probability assigned to the class label exceeds a certain threshold, initiating a new training step.
As the training steps progress, a greater number of unlabeled data is incorporated into the semi-supervised training process. Besides, the time and TF classifiers are trained together by leveraging each other's advantages through the dual domain fusion model, which transfers the integrated semantics from the previous semi-supervised training step. This approach capitalizes on the fusion of both data views to minimize prediction errors and increase the system's reliability when applied to new data.

The main reasoning behind the proposed three-model structure is that on the one hand, TF transformation is computationally expensive, but on the other hand our previous works showed that it may offer significant performance gains \cite{jalonen2023bearing_fault, alsad2024quadratic}. Hence, its corresponding model cannot operate in real-time or on edge devices and the possible performance gains cannot be directly utilized. 
Therefore, we are proposing the Dual-Domain Fusion method so that it only relies on the fast time domain model when making predictions for new samples. Nevertheless, it benefits from both representation domains in the training phase to yield the best possible performance. This technique also offers a potential deployment strategy for cloud and edge devices where the TF transformation and training of the models are solely performed on the cloud, while only the time domain model is making real-time predictions on edge devices.

In the remainder of this section, we detail each component of the proposed methodology and discuss its potential deployment strategies for cloud and edge devices.
\subsection{Time Domain Preprocessing}
Given a data acquisition system comprised of $Q$ sensors, let $\boldsymbol{s}(t) = [s_1(t), s_2(t), \cdots, s_Q(t)]^T$ be the set of measurements holding continuous stream of data, or signals, in a noisy environment, such that:
\begin{equation}
    s_q(t) = v_q(t) + \alpha_q\,\eta_q(t)\,,
\end{equation}
where $s_q(t)$ is the noisy signal or measurement recorded by sensor $q$, $v_q(t)$ is the unobserved clean measurement, $\eta_q(t)$ is the measurement's noise, and $\alpha_q$ is the noise degradation factor that defines the severity of noise 
deterioration, i.e.:
\begin{equation}
    \alpha_q = 10^{-\text{SNR}/20} \sqrt{\dfrac{\int_0^T \left|v_q(t)\right|^2\,\text{d}t}{\int_0^T \left|\eta_q(t)\right|^2\,\text{d}t}}\,\,,
\end{equation}
where $T$ is the time duration of the measurement and SNR is the signal-to-noise ratio (SNR) in decibels (dB).
\\
The acquired $Q$ measurements are preprocessed in two stages, filtering and segmentation. First, we pass the signals through a time-domain filter $h(t)$ to reduce noise and to constrain the signals' spectra to a certain band for analysis, i.e.:
\begin{equation}
    \boldsymbol{x}(t) = \left(h * \boldsymbol{s}\right)(t) = \int_0^T h(\tau)\,\boldsymbol{s}(t-\tau) \,\text{d}\tau\,,
\end{equation}
where $\tau$ denotes temporal lag. After that, we partition the filtered signals into short segments of duration $L$ with no overlap to yield manageable samples for training and testing. We express the segmented measurements as: $[\boldsymbol{x}_1(t), \boldsymbol{x}_2(t), \cdots, \boldsymbol{x}_P(t)]$,
where $\boldsymbol{x}_p(t) = [x_{1,p}(t), x_{2,p}(t), \cdots, x_{Q,p}(t)]^T$, $p$ denotes the segment index, and $P=T/L$ is the total number of segments.
\subsection{Time-Frequency Transformation}
Let $z_{q,p}(t)$ be the analytic associate of the segmented signal $x_{q,p}(t)$ obtained via the Hilbert transform, such that:
\begin{equation}
    z_{q,p}(t) = x_{q,p}(t) + j\mathcal{H}\left\{x_{q,p}(t)\right\}\,,
\end{equation}
where $\mathcal{H}\left\{\,\cdot\,\right\}$ is the Hilbert transform and $j=\sqrt{-1}$.
The time-frequency representation (TFR) of the signal, obtained by a TF distribution (TFD), illustrates the temporal evolution of its spectral content \cite{TFBook}. In other words, the TFR describes the signal's temporal and spectral information in a joint density-based representation, which reveals the signal's non-stationary dynamics that can help infer decisions about the signal under analysis \cite{9456035}.
The TFD of $z_{q,p}(t)$ is expressed by:
\begin{equation}
    \ell_{p,q}(t,f) = g_1(t) \,\underset{t}{*}\, W_{q,p}(t,f) \,\underset{f}{*}\, g_2(f)\,,
\end{equation}
\begin{equation}
    W_{q,p}(t,f) = \int z_{q,p}\left(t+\dfrac{\tau}{2}\right) z_{q,p}^*\left(t-\dfrac{\tau}{2}\right) e^{-j2\pi \tau f}\,\text{d}\tau\,,
\end{equation}
where $\ell_{p,q}(t,f)$ is the smoothed TFD of $z_{q,p}(t)$, $W_{q,p}(t,f)$ is the Wigner-Ville distribution of $z_{q,p}(t)$, $g_1(t)$ and $g_2(f)$ are separable time-domain and frequency-domain filters or kernels that control the resolution-accuracy trade-off in $\ell_{p,q}(t,f)$ \cite{9456035}, respectively, $\underset{t}{*}$ and $\underset{f}{*}$ denote the convolution operation along the temporal and spectral axes, respectively, and $z_{q,p}^*(t)$ is the complex conjugate of $z_{q,p}(t)$.
This expression can be formulated in the Doppler-lag domain, a more convenient space where convolutions become multiplications \cite{TFBook}, as:
\small
\begin{equation}
   \ell_{p,q}(t,f) = \iint G_1(\nu) \, A_{q,p}(\nu,\tau) \, G_2(\tau)\, e^{j2\pi \left(\nu t - \tau f \right)} \,\text{d}\tau\,\text{d}\nu\,,
\end{equation}
\normalsize	
\begin{equation}
    A_{q,p}(\nu,\tau) = \int z_{q,p}\left(t+\dfrac{\tau}{2}\right) z_{q,p}^*\left(t-\dfrac{\tau}{2}\right) e^{-j2\pi t \nu}\,\text{d}t\,,
\end{equation}
where $\nu$ and $\tau$ denote the Doppler and lag axes, respectively, $G_1(\nu)$ and $G_2(\tau)$ are the direct and inverse Fourier transforms of $g_1(t)$ and $g_2(f)$, respectively, and $A_{q,p}(\nu,\tau)$ is the ambiguity function of $z_{q,p}(t)$.
In this work, we employ a compact kernel distribution (CKD) to estimate the TFR of the segmented signals because of its reported high TF accuracy, resolution, and overall effectiveness \cite{9456035}. The CKD utilizes the following separable compact support kernels:
\begin{equation}
    G_1(\nu) = \exp\left(c+\frac{c\,D^2}{\nu^2-D^2}\right) : |\nu|<D\,,
\end{equation}
\begin{equation}
    G_2(\tau) = \exp\left(c+\frac{c\,E^2}{\tau^2-E^2}\right) : |\tau|<E\,,
\end{equation}
where $D\in[0,1]$ and $E\in[0,1]$ are the kernel normalized cut-offs along the Doppler and lag axes, respectively, and $c>0$ defines the shape of the kernel.
We express the TFR of the segmented measurements as: $[\boldsymbol{\ell}_1(t,f), \boldsymbol{\ell}_2(t,f), \cdots, \boldsymbol{\ell}_P(t,f)]$, where $\boldsymbol{\ell}_p(t,f) = [\ell_{1,p}(t,f), \ell_{2,p}(t,f), \cdots, \ell_{Q,p}(t,f)]^T$.
\subsection{Time and Time-Frequency Domain Classifiers} \label{sec:classifiers}
We employ conventional time domain and TF domain CNN models \cite{jalonen2023bearing_fault,alsad2024quadratic} for classification; however, our proposed semi-supervised dual-domain fusion methodology is independent of the utilized classification techniques. Therefore, it should be noted that these employed models are just one example, and the proposed framework does not rely on CNNs by any means. For instance, the models could easily be replaced with more simple machine learning methods, such as K-Nearest Neighbors, or more sophisticated deep learning methods such as large language models, or any combination of them.

The utilized models in this work are based on the same five-layer CNN structure where all convolutional layers have 64 kernels with pool size and stride set to 2. The key differences between the time domain and the TF-CNN models are their 1D and 2D convolutions, respectively, and their input sizes which are set to $(L\times f_s, Q)$ and $(M\times f_s, M\times f_s, Q)$, respectively, where $f_s$ is the sampling frequency of the data acquisition system, $M = \lambda\times L$, and $\lambda < 1$ is a down-sampling factor to reduce computational overheads. The convolutional layers in each model are followed by a dropout layer, a fully connected layer with 128 nodes, a ReLu activation function, another dropout, and a final classification layer of $N$ nodes with softmax activation where $N$ is the total number of classes to be predicted. The implementation is publicly available\textsuperscript{\ref{fn:code}} for those interested in the exact details.
\subsection{Decision Fusion Model}
Let $\mathcal{M}_{t}$ and $\mathcal{M}_{t\!f}$ be functions representing the time and the TF domain classification models, respectively. The predictions of the models can be expressed as follows:
\begin{equation}
    \widehat{\mathcal{Y}}_t = \mathcal{M}_{t}\Big(\left[\boldsymbol{x}_1(t), \boldsymbol{x}_2(t), \cdots, \boldsymbol{x}_P(t)\right]\Big)\,,
\end{equation}
\begin{equation}
    \widehat{\mathcal{Y}}_{t\!f} = \mathcal{M}_{t\!f}\Big(\left[\boldsymbol{\ell}_1(t,f), \boldsymbol{\ell}_2(t,f), \cdots, \boldsymbol{\ell}_P(t,f)\right]\Big)\,,
\end{equation}
where $\widehat{\mathcal{Y}}_t$ and $\widehat{\mathcal{Y}}_{t\!f}$ are $N\times P$ matrices holding the normalized logits from the time-domain and the TF classification models for all input segments, respectively.
We fuse the predictions of the time and TF classifiers by summing their normalized logits in a weighted fashion, i.e.:
\begin{equation}
    \widehat{\mathcal{Y}}_{\text{fused}} = \left(\boldsymbol{\beta}_{t}^T\times\mathbf{1}_P\right)\times\widehat{\mathcal{Y}}_{t} + \left(\boldsymbol{\beta}_{t\!f}^T\times\mathbf{1}_P\right)\times\widehat{\mathcal{Y}}_{t\!f}\,,
\end{equation}
where $\widehat{\mathcal{Y}}_{\text{fused}}$ holds the fused logits for all segments, $\mathbf{1}_P = [1,1,\cdots,1]\in \mathbb{R}^P$, $\boldsymbol{\beta}_{t} = [\beta_{t,1},\beta_{t,2},\cdots,\beta_{t,N}]$ and $\boldsymbol{\beta}_{t\!f} = [\beta_{t\!f,1},\beta_{t\!f,2},\cdots,\beta_{t\!f,N}]$ hold the optimal fusion weights of the time-domain and TF models, respectively, for each class.
We compute the optimal fusion weights by minimizing the sum of squared errors between the predictions and their true counterparts. Specifically, the fusion weights for class $n$ predictions are calculated as follows:
\begin{equation}
    \left[\beta_{t,n},\beta_{t\!f,n}\right]^T = 
    \left(\widehat{\mathcal{Y}}_{n} \,\widehat{\mathcal{Y}}_{n}^T\right)^{-1}\widehat{\mathcal{Y}}_{n}\,\mathcal{Y}_n^T \,,
\end{equation}
where $\widehat{\mathcal{Y}}_{n} = \left[\widehat{\mathcal{Y}}_{t,n},\widehat{\mathcal{Y}}_{t\!f,n}\right]^T$ holds the normalized logits for class $n$ from each individual model, $\mathcal{Y}$ is an $N\times P$ matrix holding the true one-hot-encoded labels for all segments, and $\mathcal{Y}_n$ is its $n$th row denoting the true binary labels for class $n$. Finally, we standardize $\widehat{\mathcal{Y}}_{\text{fused}}$ to strictly hold positive values with columns summing up to one, and compute the system's fused decisions for every input segment as follows:
\begin{equation}
    \widehat{\boldsymbol{y}} = \underset{n}{\arg\max} \Big( \widehat{\mathcal{Y}}_{\text{fused}} \Big)\,,
\end{equation}
where $\widehat{\boldsymbol{y}} = [\hat{y}_1, \hat{y}_2, \cdots, \hat{y}_P]$ hold the predicted classes that will be compared to the true ones in $\boldsymbol{y}$ to quantify performance.
%
\subsection{Confidence Thresholding}
We design a confidence thresholding mechanism to ensure that only reliable pseudo-labeled samples are included in the retraining process and to maintain class balance.
Specifically, given the set of all available samples $\mathcal{D}$, we find a set $\mathcal{S} \subseteq D$ holding samples with reliable predictions for each class $n$:
\begin{equation}
    \mathcal{S} = \left\{\mathcal{S}_1, \mathcal{S}_2, \cdots, \mathcal{S}_N\right\}\,,
\end{equation}
where $\mathcal{S}_n$ is the subset of samples holding time and TF pairs that yield high-value logits for class $n$, i.e.:
\footnotesize
\begin{equation}
    \mathcal{S}_{n} = \left\{ \left\{\boldsymbol{x}_p(t), \boldsymbol{\ell}_p(t,f)\right\} : \underset{n}{\max} \Big( \widehat{\mathcal{Y}}_{\text{fused},p} \Big) > \xi,\, \forall p \in [1,P] \right\},
\end{equation}
\normalsize
where $\xi$ is a user-defined confidence, or logits, threshold. To ensure balance between the number of samples in each class, the subsets are computed subject to the following criterion:
\begin{equation}
    \#\!\left(\mathcal{S}_{n}\right) = \dfrac{1}{N} \sum_{n=1}^N \#\!\left(\mathcal{S}_{n}\right).
\end{equation}
In other words, the number of reliable predictions is the same for each class to yield unbiased re-training data.
\subsection{Deployment Strategy for Cloud and Edge Devices} \label{sec:deployment}
The time-domain model was reported to operate in real-time even on low-cost edge devices \cite{jalonen2023bearing_fault}. This is due to its minimal memory footprint, small input size, and simple preprocessing procedures. In contrast, the TF model is known to be a non-real-time solution because of its high number of parameters, large input size, and computationally expensive data transformation process \cite{alsad2024quadratic}. Therefore, we offer a potential deployment strategy for our proposed semi-supervised dual-domain fusion method in cloud-edge networks where the training phase is solely performed in the cloud or a local server, while the real-time prediction phase is conducted on the edge device(s). This is illustrated  in Fig. \ref{fig:framework_overall}, where only a fraction of the system is deployed on the edge.

Initially, all three classifiers are trained in the cloud or the local server in a supervised manner by holding out a testing set. After that, a copy of the trained time-domain classifier is sent out to all edge devices for real-time inference. We keep a copy of this trained classifier in the cloud and refer to it as the original time-domain model. Then, edge devices send back the new unlabeled samples to the cloud or server to conduct the proposed semi-supervised dual-domain fusion procedure. Note that this procedure does not require fast communication and can be initiated whenever a sufficient amount of unlabeled data is received. We compare the newly updated fusion-trained time-domain model to its original version using the held-out testing set. If we detect a considerable performance gain, the updated fusion-trained time-domain model is sent out to replace its original version.
Additionally, the fused non-real-time model in the cloud or server can be used to correct the prediction log on each edge device. However, the continuous dependence on this service can be decided by the end-user. In other words, the end-user has three options: (1) operate offline by utilizing the originally trained fast model with no updates, (2) permit scheduled updates for the fast time-domain model from the server, and (3) permit full support from both the edge device and the cloud service which includes: real-time monitoring, scheduled model updates for higher performance, and fault record correction via the fused non-real-time model for better reliability. This proposed framework enables faster fault detection, more reliable predictions, lower energy consumption which makes battery operation feasible, and lower hardware costs compared to using one large model either in the cloud or on the edge device. In brief, our proposed method and its deployment strategy can fuse the best parts of the two domains: the high accuracy from the TF-domain model and the faster predictions from the time-domain model.
\section{Experimental Validations} \label{sec:experimental_validations}
The validity of our method is tested by applying it to the problem of bearing fault diagnosis under time-varying speeds with and without added noise. The overview of the proposed method is shown in Fig. \ref{fig:framework_overall}. Specifically, we conduct two case studies to determine if the proposed Dual-Domain Fusion method can yield consistent reliable predictions across different variables; see Table \ref{table:parameters}. Besides we compare it to five other semi-supervised training techniques to quantify its possible gains in performance and robustness.
\subsection{Case Studies and Preprocessing}
\subsubsection{Case study KAIST}
This experiment uses vibration signals to diagnose bearing faults under rapidly varying motor speeds. We used an open-access dataset from the Korean Advanced Institute of Science and Technology (KAIST) \cite{KAIST_data}, which includes four classes: \emph{Normal}, \emph{Outer}, \emph{Inner}, and \emph{Ball}, describing a typical bearing function and three common faults. For each of these classes, 35 minutes of vibration data were acquired using two accelerometers installed on the x- and y-axes of the bearing housing.
The signals were originally sampled at 25.6 kHz, but we downsampled them to 20 kHz using a Finite Impulse Response (FIR) anti-aliasing lowpass filter with delay compensation. Additionally, we partitioned the filtered signals into 100 ms segments and added white Gaussian noise at -5 dB signal-to-noise ratio
(SNR) while also retaining the clean data. Finally, we computed the TFR of the segmented measurements using the CKD and downsampled their TF representations. Finally, we selected the first 1300 segments from each class to allow us to conduct an extensive hyperparameter search with reasonable computing time.
\subsubsection{Case study SQV}
This experiment includes diagnosing the severity of bearing faults under well-defined time-varying speeds using vibration signals. We used the publicly available Spectra Quest Vibration (SQV) dataset which includes seven classes: \emph{Normal}, \emph{Outer 1}, \emph{Outer 2}, \emph{Outer 3}, \emph{Inner 1}, \emph{Inner 2}, and \emph{Inner 3} \cite{liu2022subspace}. This dataset describes a typical bearing function and two common faults with three degrees of severity: 1 is mild, 2 is moderate, and 3 indicates a severe fault condition. For each of these classes, 2.7 minutes, on average, of vibration data were acquired using one accelerometer. These measurements were made under a continuously varying motor speed set to increase from rest to 3000 rpm, stay at 3000 rpm for a while, and then continuously decelerate to rest again.
We linearly interpolated the measured speed to match the temporal sampling of the vibration signals at 25.6 kHz. Besides, we automatically extracted regions of interest corresponding to periods when the speed was significantly non-zero (on average above 200 rpm). This process resulted in reducing the duration of the vibration signals to 1.84 minutes on average. After that, we partitioned the vibration signals into 78.125 ms segments and added white Gaussian noise at -5 dB signal-to-noise ratio (SNR) while also retaining the clean data. Finally, we computed and downsampled the TFR of the remaining segments. There was a class imbalance problem, which we solved by selecting the first 1300 segments from each class, similarly to the first case study.
\begin{table}[!t]
\footnotesize
\centering
\caption{The pre-processing parameters for the two case studies.}
\begin{tabular}{ccc}
\toprule
\textbf{Parameters} & \textbf{Case Study KAIST} & \textbf{Case Study SQV}
\\\midrule
$N$ & $4$ classes & $7$ classes 
\\\midrule
$Q$ & $2$ sensors & $1$ sensor
\\\midrule
$T$ & \begin{tabular}{@{}c@{}}$35$ min per\\ sensor per class\end{tabular} & \begin{tabular}{@{}c@{}}$\approx2.7$ min per\\ sensor per class\end{tabular}
\\\midrule
$L$ & $100$ ms & $78.125$ ms
\\\midrule
$P$ & \begin{tabular}{@{}c@{}}$21,000$ segments\\ per sensor per class\end{tabular} & \begin{tabular}{@{}c@{}}$\approx 1,407$ segments\\ per sensor per class\end{tabular}
\\\midrule
$f_s$ & $20$ kHz & $25.6$ kHz
\\\midrule
$h(t)$ & \begin{tabular}{@{}c@{}}FIR lowpass filter\\with delay compensation\end{tabular} & ---
\\\midrule
SNR & \{$-5$ dB, Clean\} & \{$-5$ dB, Clean\}
\\\midrule
$\lambda$ & $0.064$ & $0.064$
\\\midrule
$\{c,D,E\}$ & $\{1,0.1,0.1\}$ & $\{1,0.1,0.1\}$
\\\bottomrule
\label{table:parameters}
\end{tabular}
\end{table}
\subsection{Experimental Setups}
\subsubsection{Temporal Data Splits}
We divided the preprocessed datasets into ten equal temporal splits. The first split (Split 1), which is 10\% of all data, has ground truth labels. The ground truth labels of the next six splits (Splits 2–7) were removed. Together these first seven splits (Splits 1-7) form the training data. The next split (Split 8), was set aside as the validation set used for parameter search, and the last two splits (Splits 9–10) were reserved for testing. 
\subsubsection{Iterative training Steps}
The training is done iteratively by adding one Split at a time, meaning there are seven training steps (Steps 1-7) in total. Step 1 is the initial supervised training step. This is also illustrated on the left in Fig.\ref{fig:framework_overall}. FixMatch, MixMatch and Adversarial training require both labeled and unlabeled data so we skipped Step 1 for them and started from Step 2. We acquired results for both clean and -5 dB conditions without changing any parameters.
\subsubsection{Comparisons}
We compared our Dual-Domain Fusion method against five well-known semi-supervised benchmarks: FixMatch \cite{sohn2020fixmatch}, MixMatch \cite{berthelot2019mixmatch}, Mean Teacher \cite{tarvainen2017mean}, Adversarial training, and Self-training. All the methods, including ours, were tested using the same time-domain classifier (details in section \ref{sec:classifiers}) and data. This means there was no computational complexity difference during testing inference or on an edge device as explained in section \ref{sec:deployment}. Thus, testing performance gains or losses are solely based on the different training strategies of each method.


\subsubsection{Hyperparameter search}
We trained our Dual-Domain Fusion method with different confidence thresholds $\xi$; ranging from 0.0 to 0.9 with 0.1 increment. We used all the clean training data (Step 7), trained for three repetitions, and selected thresholds that maximized validation accuracy.

We also conducted an exhaustive grid search to find the best combination of hyperparameters for FixMatch, MixMatch, Mean-Teacher, and Adversarial training for each case study. These were trained only once to keep the computational time reasonable. The grid search had 4 variables each with 2 options yielding 16 different settings. While the epoch number was smaller than the warm-up epochs hyperparameter, the consistency weight of Mean Teacher and the unsupervised weight of the other methods were set to 0.0. Note, that the original implementation of FixMatch uses the confidence threshold $\xi$ of 0.95, but we also trained it with 0.8.

\begin{table}[!t]
\footnotesize
\centering
\caption{Validation accuracies averaged over three repetitions $\pm$ standard deviations of our Dual-Domain Fusion method with different confidence thresholds.}
\begin{tabular}{ccc}
\toprule
\textbf{Confidence Threshold} & \textbf{Case Study KAIST} & \textbf{Case Study SQV}
\\\midrule
0.0 & $88.08 \pm 2.44$ & $85.27 \pm 3.26$
\\\midrule
0.1 & $87.95 \pm 1.76$ & $86.01 \pm 1.94$
\\\midrule
0.2 & $88.21 \pm 1.41$ & $83.63 \pm 4.35$
\\\midrule
0.3 & $88.78 \pm 0.79$ & $\mathbf{88.24 \pm 0.93}$
\\\midrule
0.4 & $88.33 \pm 1.18$ & $84.91 \pm 6.17$
\\\midrule
0.5 & $\mathbf{90.06 \pm 1.98}$ & $87.91 \pm 1.06$
\\\midrule
0.6 & $86.79 \pm 0.48$ & $85.75 \pm 1.50$
\\\midrule
0.7 & $87.18 \pm 0.33$ & $84.84 \pm 2.56$
\\\midrule
0.8 & $86.54 \pm 0.57$ & $86.01 \pm 1.98$
\\\midrule
0.9 & $88.27 \pm 2.88$ & $83.48 \pm 1.14$
\\\bottomrule
\label{table:DDF_validation_results}
\end{tabular}
\end{table}
%
%
\begin{table*}[!t]
\footnotesize
\centering
\caption{The hyperparameter search space and results for comparisons. The best hyperparameter combinations are shown in bold and the corresponding best validation accuracy scores are in the bottom row.}
\begin{tabular}{ccccccccc}
\toprule
Case Study & \multicolumn{4}{c}{KAIST} & \multicolumn{4}{c}{SQV}
\\\midrule
& \textbf{FixMatch} & \textbf{MixMatch} & \textbf{\begin{tabular}{@{}c@{}}Mean\\teacher\end{tabular}} & \textbf{\begin{tabular}{@{}c@{}}Adversarial\\training\end{tabular}} & \textbf{FixMatch} & \textbf{MixMatch} & \textbf{\begin{tabular}{@{}c@{}}Mean\\teacher\end{tabular}} & \textbf{\begin{tabular}{@{}c@{}}Adversarial\\training\end{tabular}}
\\\midrule
Learning rate 
& $10^{-4}, \mathbf{10^{-5}}$ 
& $10^{-4}, \mathbf{10^{-5}}$ 
& $\mathbf{10^{-4}}, 10^{-5}$ 
& $10^{-4}, \mathbf{10^{-5}}$ 
& $\mathbf{10^{-4}}, 10^{-5}$ 
& $10^{-4}, \mathbf{10^{-5}}$ 
& $10^{-4}, \mathbf{10^{-5}}$ 
& $10^{-4}, \mathbf{10^{-5}}$
\\\midrule
Warm-up epochs 
& $\mathbf{0}, 20$ 
& $0, \mathbf{20}$ 
& $0, \mathbf{20}$ 
& $0, \mathbf{20}$ 
& $\mathbf{0}, 20$ 
& $\mathbf{0}, 20$ 
& $\mathbf{0}, 20$ 
& $0, \mathbf{20}$
\\\midrule
Unsupervised weight 
& $0.5, \mathbf{1.0}$ 
& $\mathbf{0.5}, 1.0$ 
& --- 
& $0.5, \mathbf{1.0}$
& $0.5, \mathbf{1.0}$ 
& $\mathbf{0.5}, 1.0$ 
& --- 
& $\mathbf{0.5}, 1.0$
\\\midrule
Confidence threshold 
& $\mathbf{0.80}, 0.95$ 
& --- 
& --- 
& --- 
& $\mathbf{0.80}, 0.95$ 
& --- 
& --- 
& ---
\\\midrule
Sharpening temperature 
& --- 
& $\mathbf{0.2}, 0.5$ 
& --- 
& --- 
& --- 
& $0.2, \mathbf{0.5}$ 
&  --- 
& ---
\\\midrule
Consistency weight 
& --- 
& --- 
& $\mathbf{1.0}, 10.0$ 
& ---
& --- 
& --- 
& $\mathbf{1.0}, 10.0$ 
& ---
\\\midrule
EMA decay 
& --- 
& --- 
& $\mathbf{0.99}, 0.999$ 
& ---
& --- 
& --- 
& $0.99, \mathbf{0.999}$ 
& ---
\\\midrule
Attack magnitude 
& --- 
& --- 
& --- 
& $\mathbf{0.2}, 0.3$
& --- 
& --- 
& --- 
& $\mathbf{0.2}, 0.3$
\\\midrule
\textbf{Validation Accuracy}
& $67.69$ & $68.85$ & $57.88$ & $66.35$
& $58.90$ & $57.36$ & $59.78$ & $57.36$
\\\bottomrule
\label{table:hyperparameter_results}
\end{tabular}
\end{table*}
\subsection{Results}
The validation set results of our Dual-Domain Fusion method are shown in Table \ref{table:DDF_validation_results}. The hyperparameter search of different confidence thresholds reveals that 0.5 / 0.3 yields the highest validation accuracy of 90\% / 88\% on the case study KAIST / SQV. Confidence thresholding increased the accuracies by 2 (KAIST) and 3 (SQV) percentage points, as depicted from the results with 0.0 confidence threshold.

The hyperparameter search space, best parameter combinations and best validation accuracies of FixMatch, MixMatch, Mean teacher and Adversarial training are listed in Table \ref{table:hyperparameter_results}. The methods reached consistent results, 66-69\% on KAIST and 57-60\% on SQV, only exception being Mean teacher remaining at 58\% on KAIST. These are significantly lower scores compared to the Dual-Domain Fusion method. Three out of eight of the best hyperparameters did not vary over the case studies, although only one method was using each of those three. This suggests the searched hyperparameters and their values were reasonable.

Table \ref{table:tab_perf_results} presents the testing performance of the proposed Dual-Domain Fusion method and compares it to the well known semi-supervised benchmarks. The table summarizes the testing accuracy for the two case studies, for clean and -5dB noise level data, and across the increasing number of training samples. On the one hand, one notes that the accuracy of all methods increase when moving from -5 dB SNR to clean data. On the other hand, by examining the trends across the increasing number of unlabeled training samples, we find that the testing scores for the benchmark methods are practically unchanged, only exception being FixMatch on the clean SQV data. At the same time the accuracy of the proposed Dual-Domain Fusion method increases remarkably. In fact, Dual Domain Fusion achieves significantly higher performance than the comparative methods in both case studies and noise levels when most of the training data is available. For instance, it reaches 88\% (KAIST) and 94\% (SQV) accuracy when supplied with clean signals compared to only 80\% and 81\% for the second best Self-training method. However, Self-training is performing slightly better at Steps 2-3 on the clean KAIST data while Dual-Domain Fusion is experiencing a performance dip, but Dual Domain Fusion still surpasses it by 8\% at the final Step 7. The severe noise level of -5 dB is too challenging for MixMatch, Mean teacher and Adversarial training on KAIST as they reach only 25-26\% in the 4-class classification problem. The same phenomenon is present with Fixmatch on SQV as its accuracy is only 16\% in the 7-class problem despite the extensive hyperparameter search.

\begin{figure}[!t]
    \centering
    \subfloat[Testing performance using the KAIST dataset. \label{fig:barplot_accuracies_KAIST}]{\includegraphics[width=.475\textwidth]{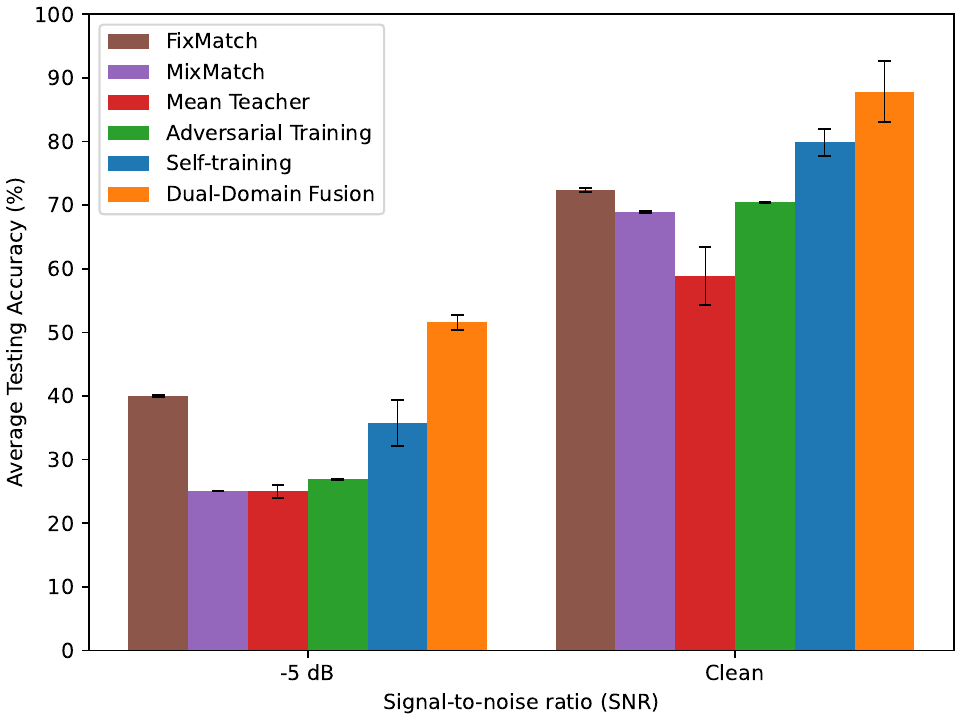}}
    \\
    \subfloat[Testing performance using the SQV dataset. \label{fig:barplot_accuracies_SQV}]{\includegraphics[width=.475\textwidth]{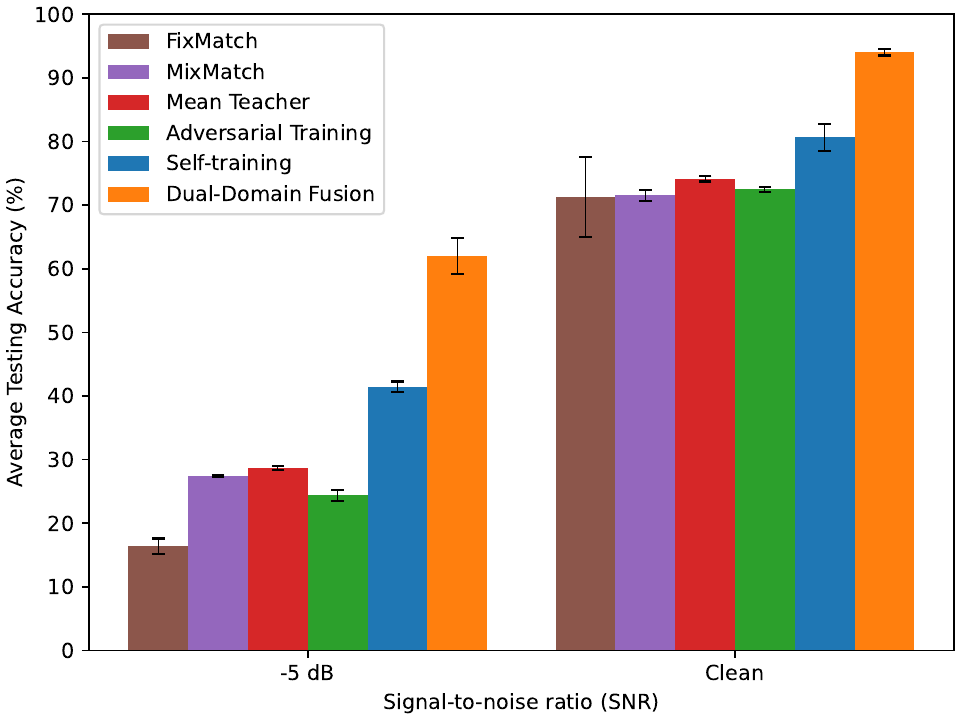}}
    \caption{The testing accuracies of the proposed Dual-Domain fusion and comparison techniques. Averaged results are depicted across the different SNR levels along with their corresponding 95\% confidence intervals and relative gains in performance.}
    \label{fig:barplot_accuracies}
\end{figure}

\begin{figure}[!t]
    \centering
    \subfloat[Case study: KAIST. \label{fig:acc_std_clean_KAIST}]{\includegraphics[width=.475\textwidth]{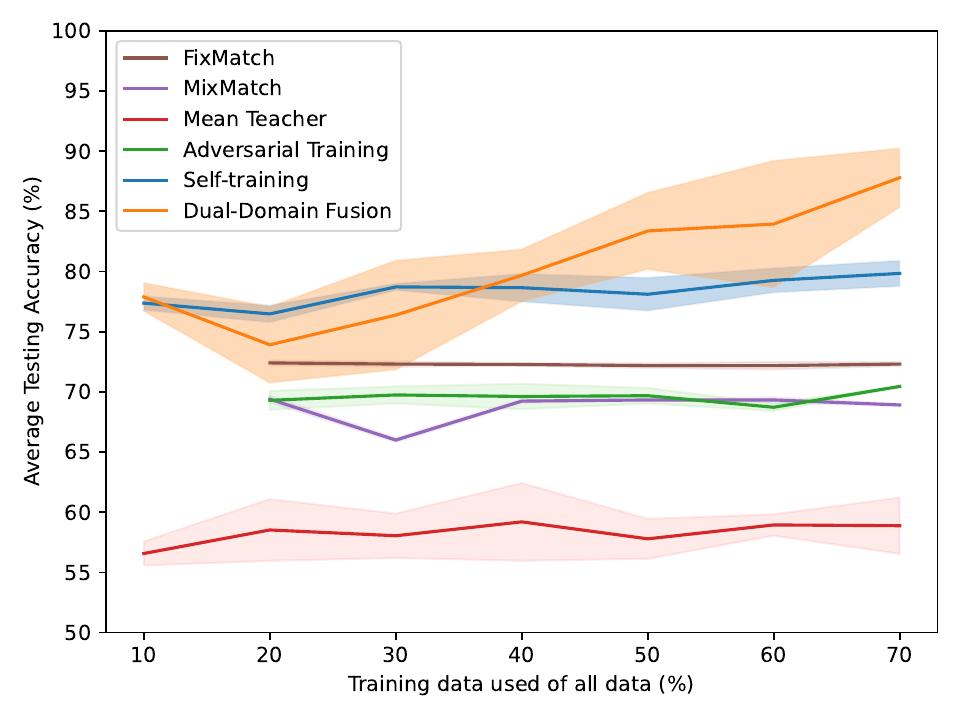}}
    \\
    \subfloat[Case study: SQV. \label{fig:acc_std_clean_SQV}]{\includegraphics[width=.475\textwidth]{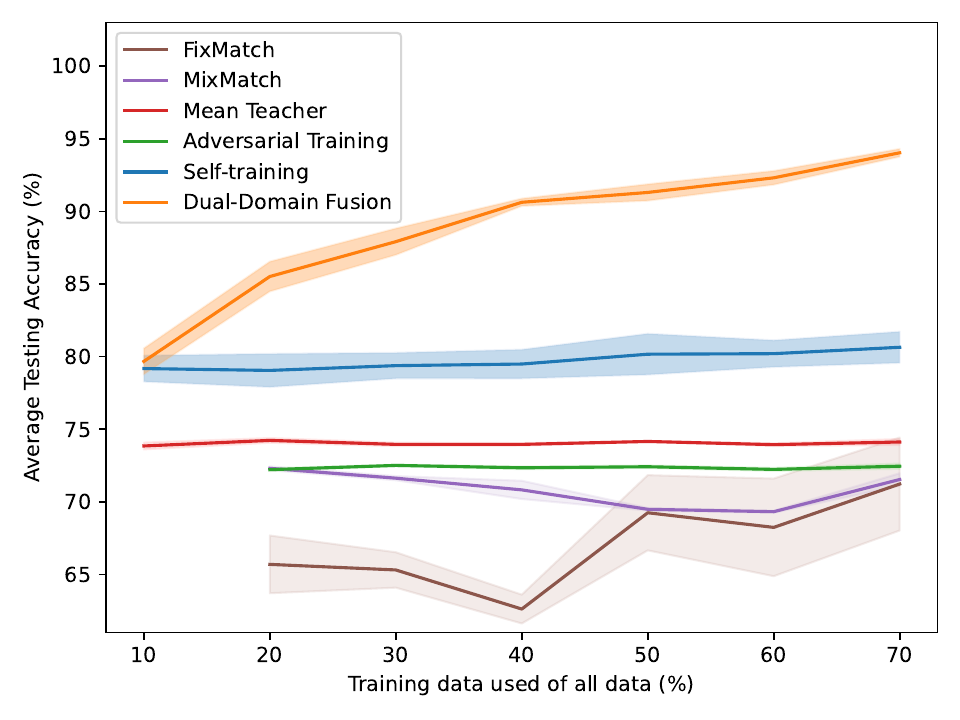}}
    \caption{The relationship between the number of training samples and the testing performance for our Dual-Domain Fusion and comparison methods. The results show the averaged testing accuracy $\pm$ standard deviation when utilizing the clean data in each case study.}
    \label{fig:acc_std_clean}
\end{figure}

Figs. \ref{fig:barplot_accuracies} and \ref{fig:acc_std_clean} further examine the gain in performance delivered by the proposed Dual-Domain Fusion method.
In Fig. \ref{fig:barplot_accuracies}, the testing accuracy, when using all the unlabeled training data, is illustrated with bar plots across the different noise levels and for the two case studies. The results also demonstrate the accuracy's 95\% confidence interval and the relative gains delivered by our proposed method. By inspecting the accuracy bars, it becomes apparent that our Dual-Domain Fusion method outperforms the competing techniques across both noise levels.

\begin{figure*}[!t]
    \centering
    \subfloat[FixMatch (KAIST)]{%
        \includegraphics[width=0.32\textwidth]{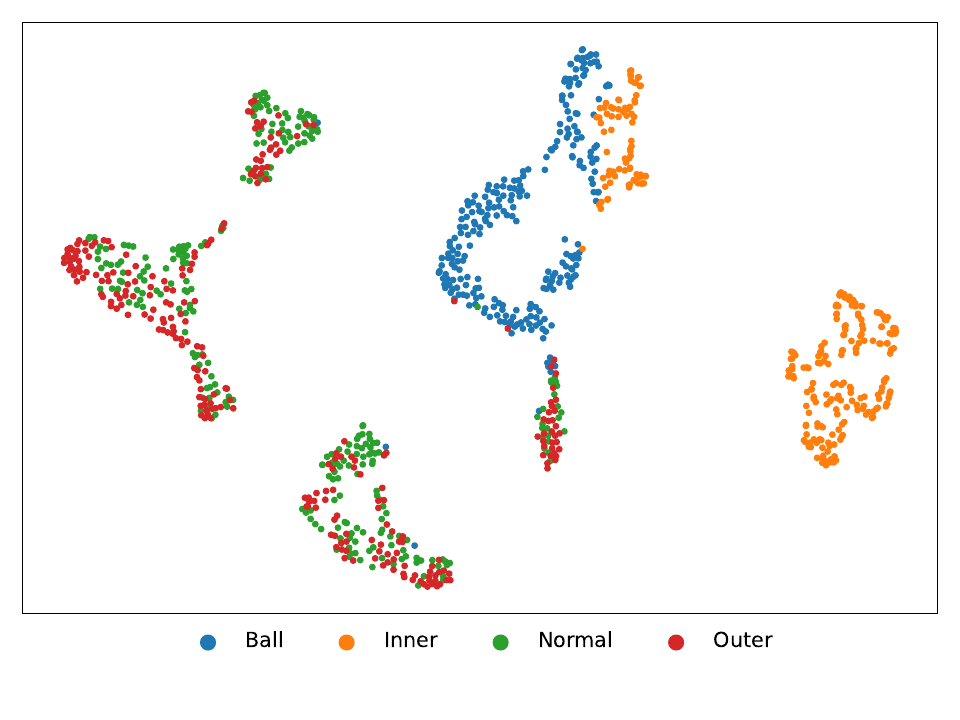}
    }\hfill
    \subfloat[MixMatch (KAIST)]{%
        \includegraphics[width=0.32\textwidth]{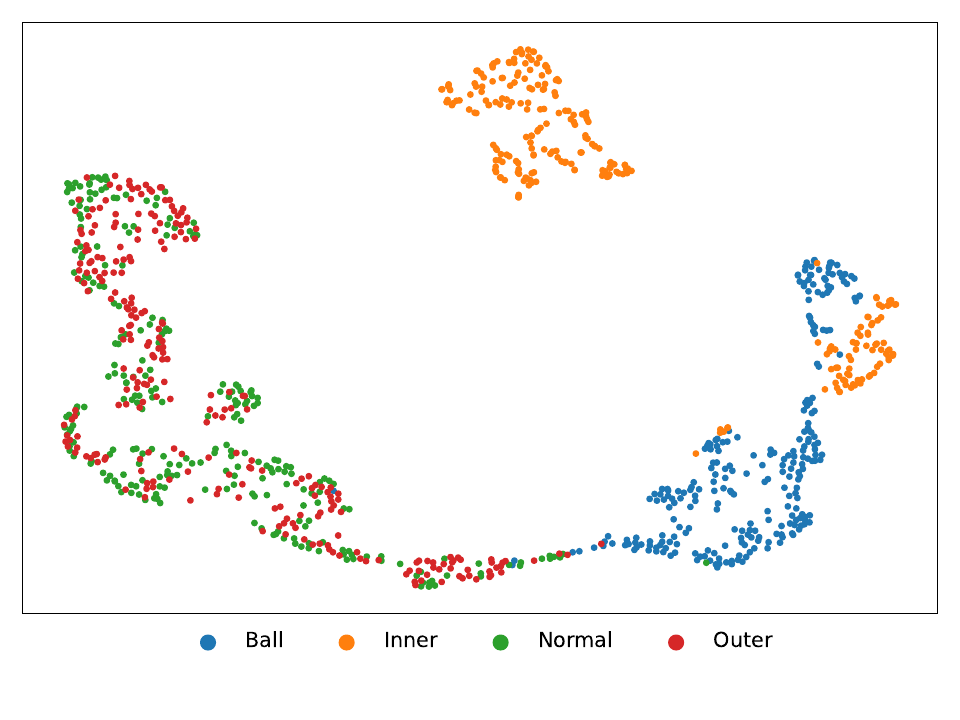}
    }\hfill
    \subfloat[Mean Teacher (KAIST)]{%
        \includegraphics[width=0.32\textwidth]{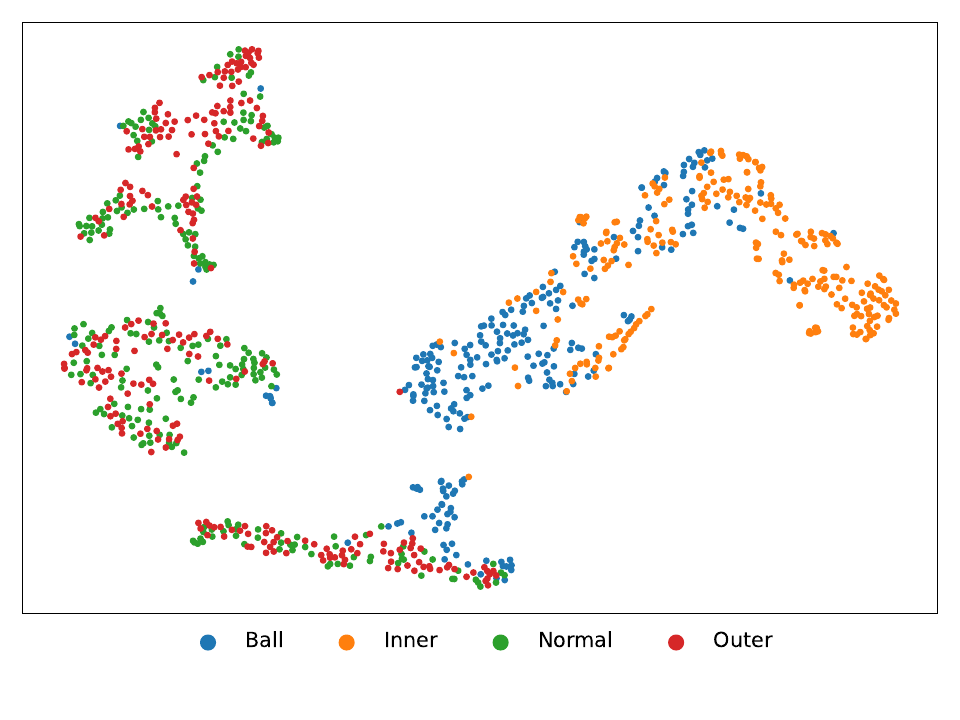}
    }

    \vspace{0.3cm} 

    \subfloat[Adversarial training (KAIST)]{%
        \includegraphics[width=0.32\textwidth]{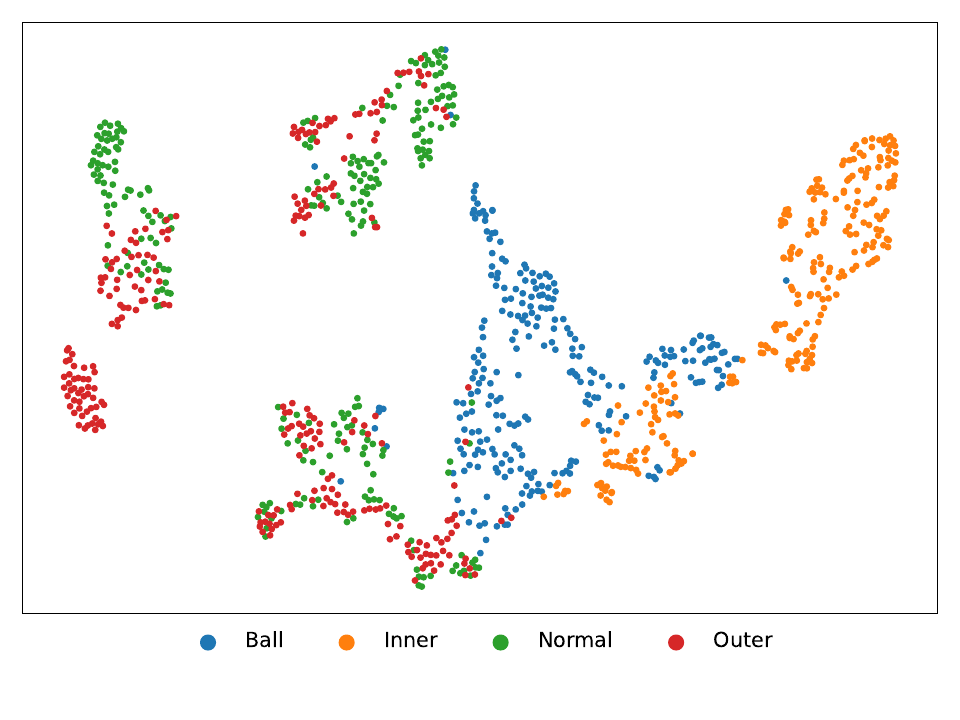}
    }\hfill
    \subfloat[Self-training (KAIST)]{%
        \includegraphics[width=0.32\textwidth]{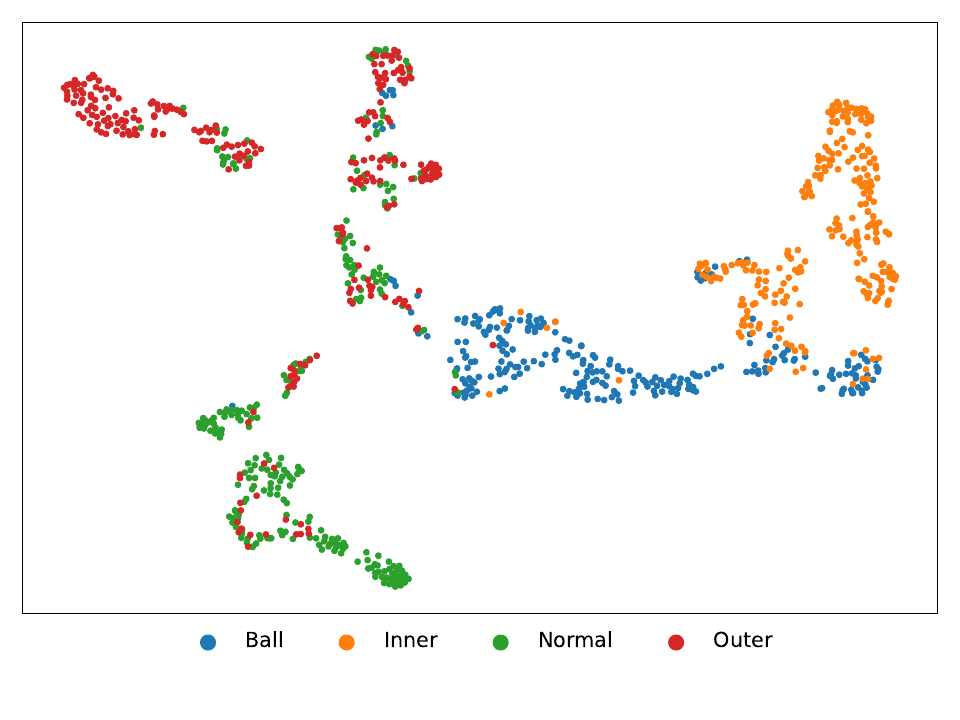}
    }\hfill
    \subfloat[Dual-Domain Fusion (KAIST)]{%
        \includegraphics[width=0.32\textwidth]{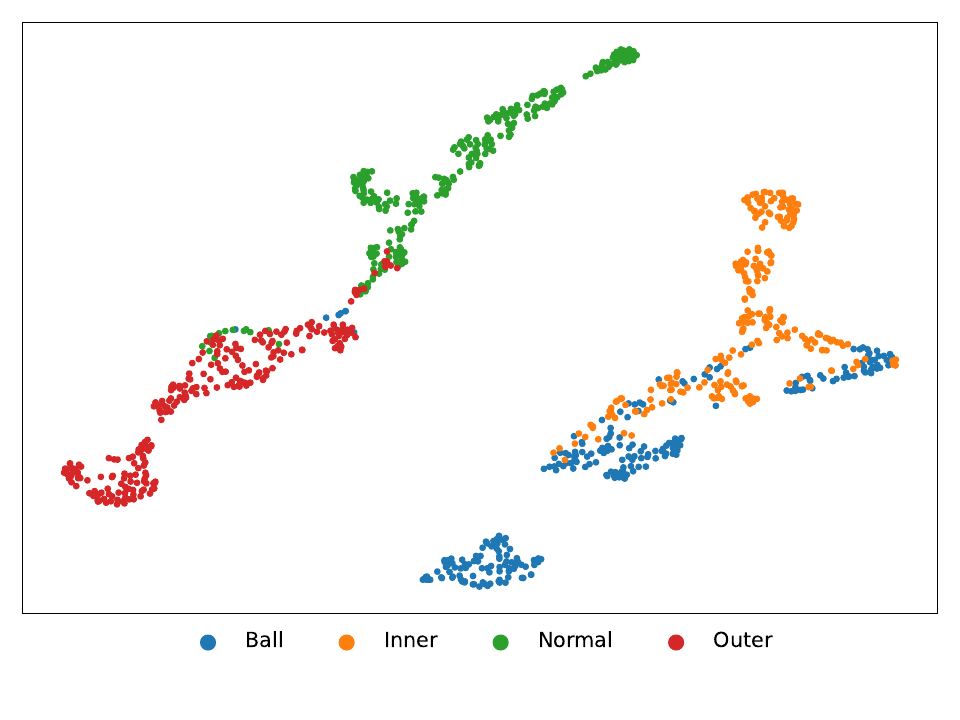}
    }
    
    \caption{T-SNE results of the KAIST case study using clean data.}
    \label{fig:t-sne_KAIST}
\end{figure*}

\begin{figure*}[!t]
    \centering
    \subfloat[FixMatch (SQV)]{%
        \includegraphics[width=0.32\textwidth]{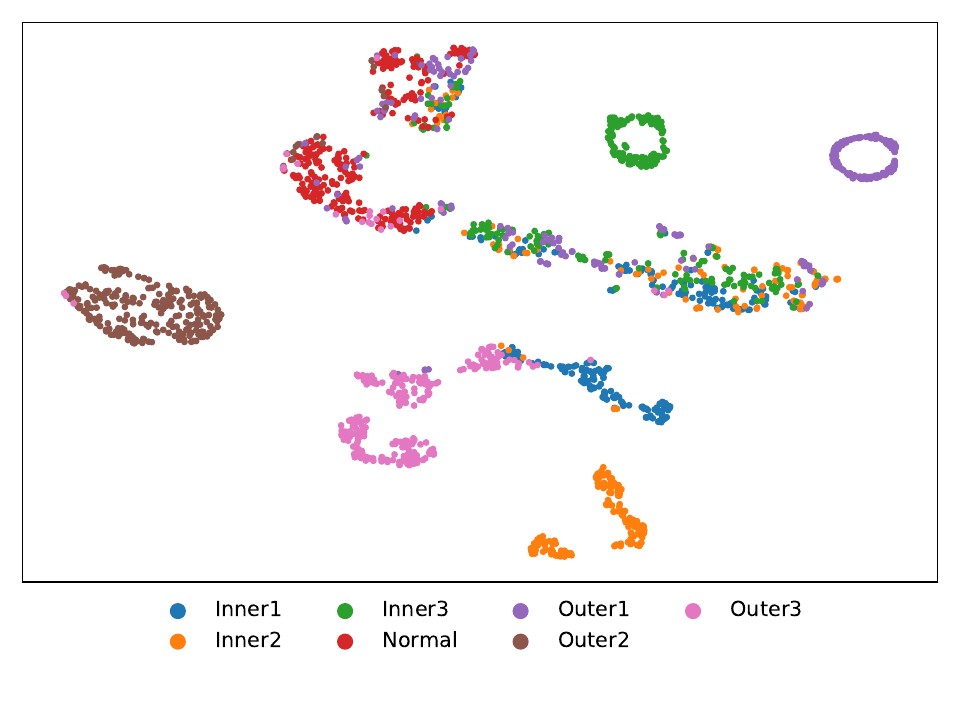}
    }\hfill
    \subfloat[MixMatch (SQV)]{%
        \includegraphics[width=0.32\textwidth]{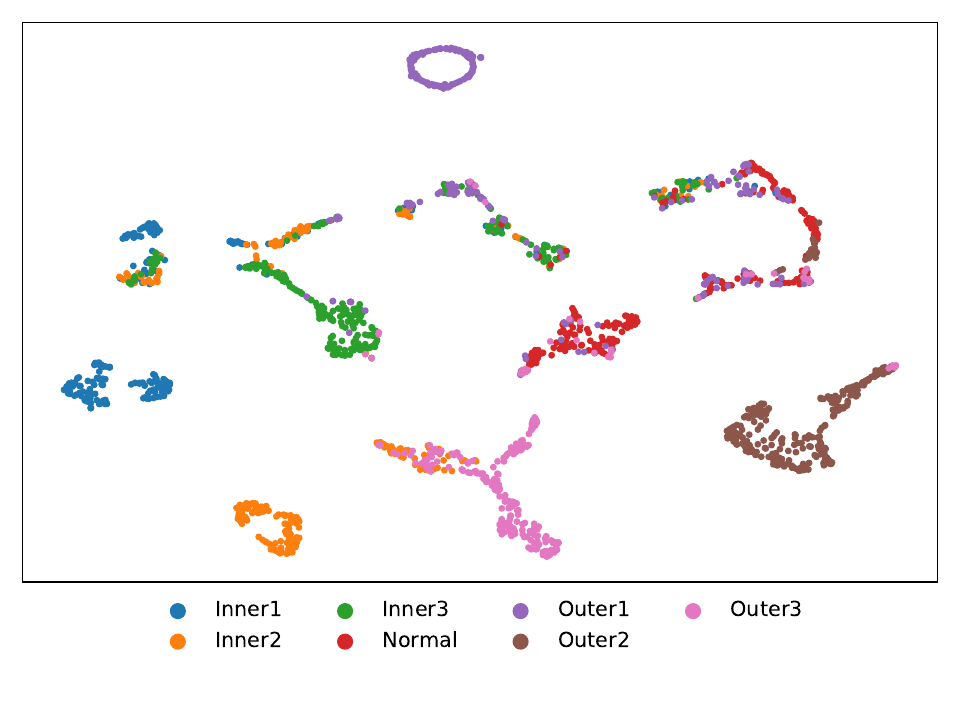}
    }\hfill
    \subfloat[Mean Teacher (SQV)]{%
        \includegraphics[width=0.32\textwidth]{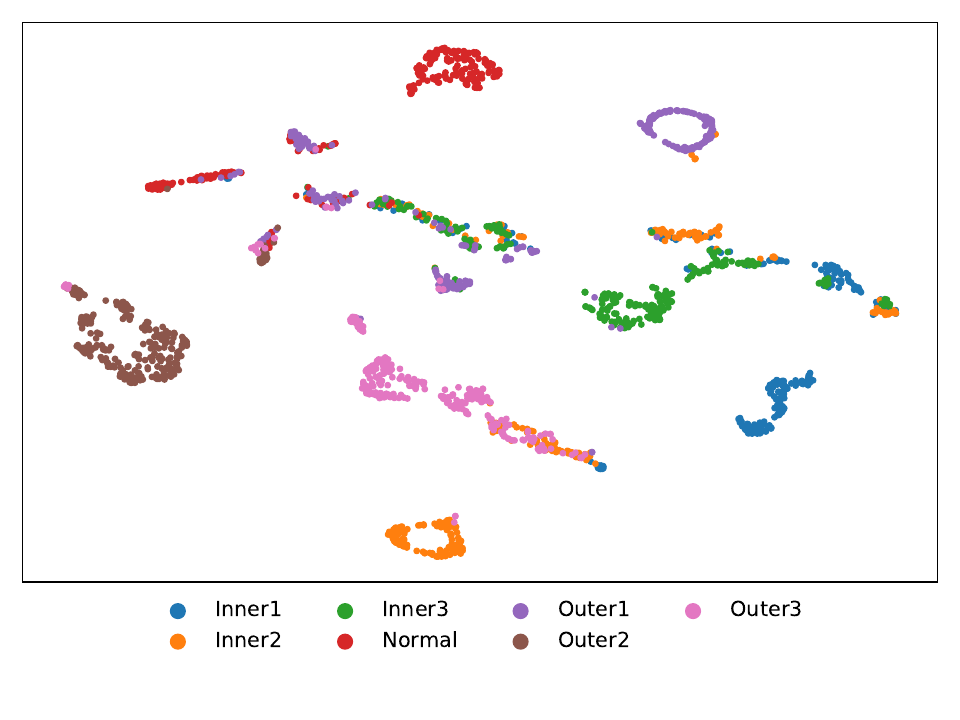}
    }

    \vspace{0.3cm} 

    \subfloat[Adversarial training (SQV)]{%
        \includegraphics[width=0.32\textwidth]{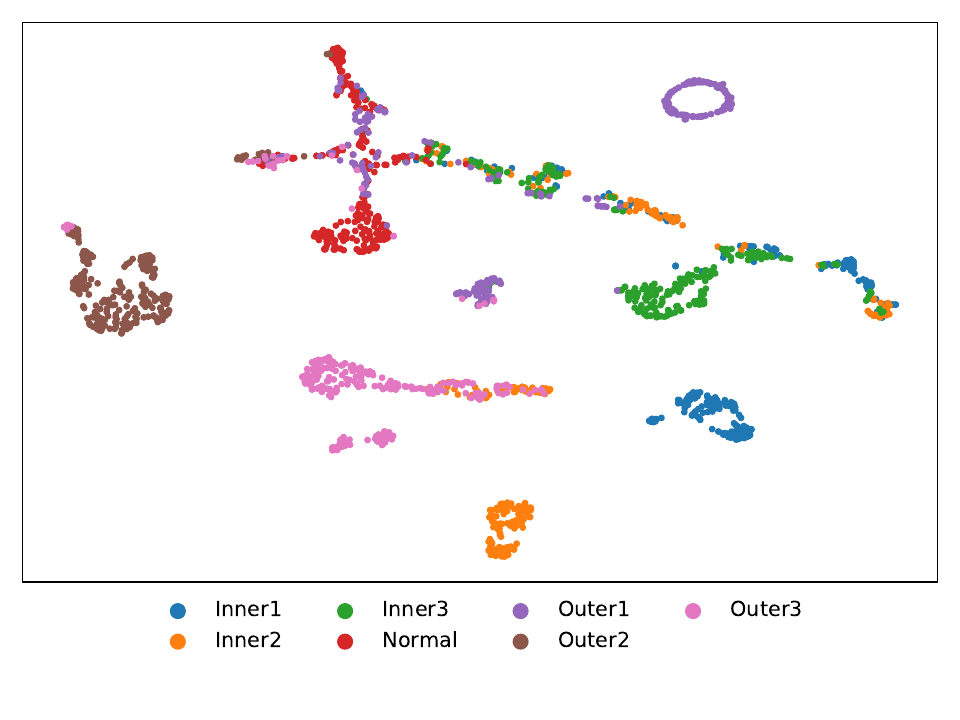}
    }\hfill
    \subfloat[Self-training (SQV)]{%
        \includegraphics[width=0.32\textwidth]{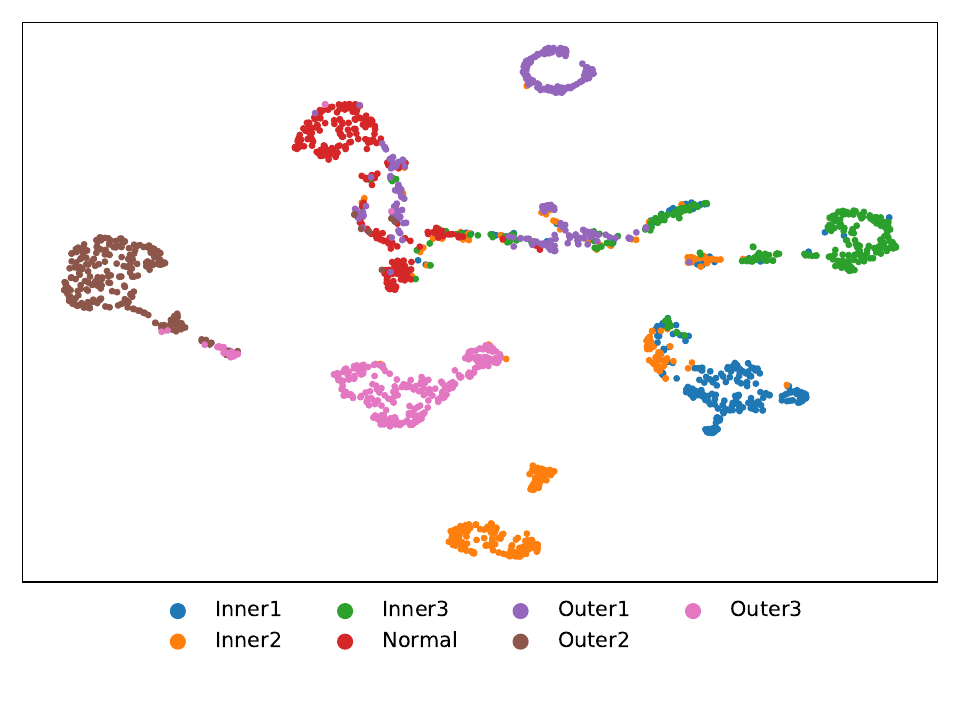}
    }\hfill
    \subfloat[Dual-Domain Fusion (SQV)]{%
        \includegraphics[width=0.32\textwidth]{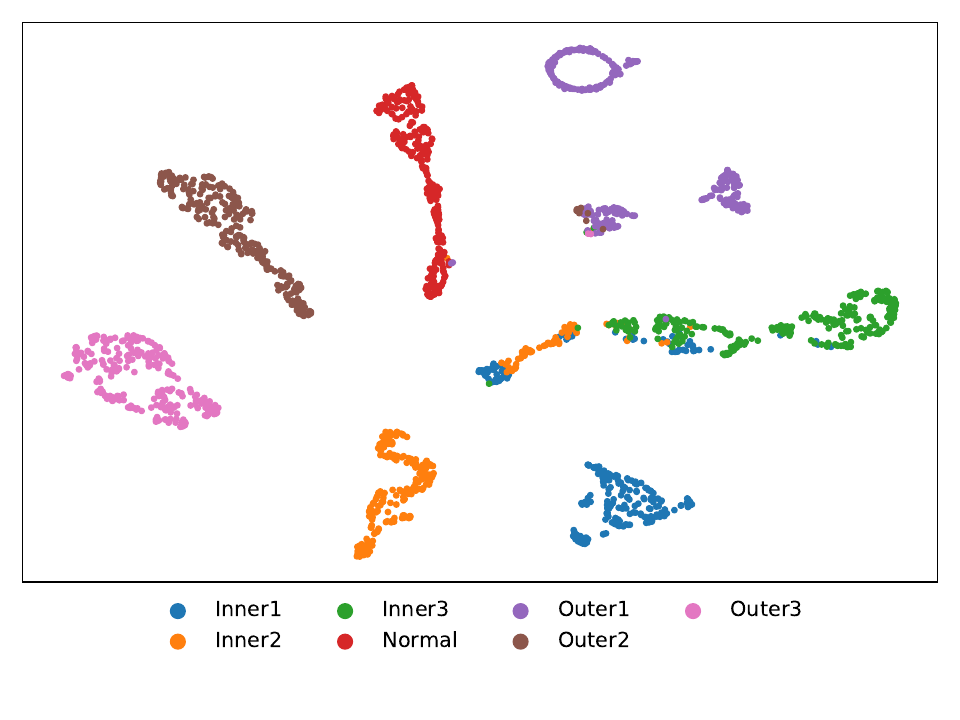}
    }
    
    \caption{T-SNE results of the SQV case study using clean data.}
    \label{fig:t-sne_SQV}
\end{figure*}

Moreover, Fig. \ref{fig:acc_std_clean} expands this comparison by displaying the relationship between the averaged testing accuracy and the number of training samples for the clean case. The reported trends reveal that the comparison techniques did not benefit from the increase in the number of unlabeled training samples most of the time. Specifically, they demonstrate a statistically flat performance profile for the KAIST and SQV datasets; see Figs. \ref{fig:acc_std_clean_KAIST} and \ref{fig:acc_std_clean_SQV}. With the exception of FixMatch in SQV, our proposed DDF method offers major gains in performance that are positively correlated with the amount of supplied data. In other words, it achieves higher accuracy given more unlabeled training samples. Furthermore, the results show an upward trend for the DDF performance in both case studies. This implies that the final accuracy could be even higher if more unlabeled training data was available. Nonetheless, the standard deviation of the DDF is the highest on KAIST.
\begin{table*}[!t]
\centering
\caption{The testing set accuracy scores averaged over three repetitions $\pm$ standard deviations. The best results are highlighted in bold across the two noise levels, training steps, and case studies.}
\begin{adjustbox}{max width=\textwidth}
\begin{tabular}{ccccccccc}
\toprule
Case Study & Training data (\%) & Noise level (dB) & FixMatch & MixMatch & Mean Teacher & Adversarial Training & Self-Training & Dual-Domain Fusion
\\\midrule
\multirow{15}{*}{KAIST}
& 10 & -5 dB
& ---
& ---
& $28.14 \pm 4.35$ 
& --- 
& $37.79 \pm 0.54$
& $\mathbf{38.78 \pm 0.96}$
\\
& 20 & -5 dB 
& $40.10 \pm 0.16$ 
& $40.90 \pm 0.12$ 
& $26.79 \pm 0.55$ 
& $26.92 \pm 0.00$ 
& $36.06 \pm 0.44$
& $\mathbf{41.06 \pm 4.48}$
\\
& 30 & -5 dB 
& $40.29 \pm 0.14$ 
& $29.74 \pm 0.12$ 
& $25.00 \pm 1.36$ 
& $26.92 \pm 0.00$
& $35.67 \pm 1.88$
& $\mathbf{46.57 \pm 1.54}$
\\
& 40 & -5 dB 
& $39.87 \pm 0.05$ 
& $25.00 \pm 0.00$ 
& $27.47 \pm 2.48$ 
& $26.92 \pm 0.00$
& $35.90 \pm 1.00$
& $\mathbf{47.15 \pm 0.69}$
\\
& 50 & -5 dB 
& $39.87 \pm 0.20$ 
& $31.31 \pm 0.12$ 
& $25.87 \pm 1.10$ 
& $26.92 \pm 0.00$
& $34.39 \pm 0.65$
& $\mathbf{48.40 \pm 1.42}$
\\
& 60 & -5 dB 
& $39.97 \pm 0.05$ 
& $25.00 \pm 0.00$ 
& $26.60 \pm 1.32$ 
& $26.92 \pm 0.00$
& $35.48 \pm 1.93$
& $\mathbf{51.60 \pm 1.01}$
\\
& 70 & -5 dB 
& $40.00 \pm 0.08$ 
& $25.00 \pm 0.00$ 
& $25.03 \pm 0.51$ 
& $26.89 \pm 0.05$ 
& $35.77 \pm 1.84$
& $\mathbf{51.57 \pm 0.59}$
\\
\\
& 10 & Clean
& --- 
& --- 
& $56.57 \pm 1.00$ 
& --- 
& $77.37 \pm 0.58$
& $\mathbf{77.88 \pm 1.17}$
\\
& 20 & Clean
& $72.40 \pm 0.21$ 
& $69.39 \pm 0.28$ 
& $58.53 \pm 2.56$ 
& $69.29 \pm 0.79$ 
& $\mathbf{76.47 \pm 0.70}$
& $73.91 \pm 3.16$
\\
& 30 & Clean
& $72.31 \pm 0.21$ 
& $65.99 \pm 0.16$ 
& $58.04 \pm 1.84$ 
& $69.74 \pm 0.70$ 
& $\mathbf{78.72 \pm 0.28}$
& $76.38 \pm 4.54$
\\
& 40 & Clean
& $72.28 \pm 0.12$ 
& $69.23 \pm 0.16$ 
& $59.20 \pm 3.22$ 
& $69.62 \pm 1.06$ 
& $78.65 \pm 1.16$
& $\mathbf{79.67 \pm 2.16}$
\\
& 50 & Clean
& $72.18 \pm 0.18$  
& $69.33 \pm 0.00$ 
& $57.79 \pm 1.65$ 
& $69.68 \pm 0.64$ 
& $78.11 \pm 1.37$
& $\mathbf{83.37 \pm 3.19}$
\\
& 60 & Clean
& $72.18 \pm 0.32$ 
& $69.33 \pm 0.21$ 
& $58.94 \pm 0.88$ 
& $68.72 \pm 0.30$ 
& $79.26 \pm 1.01$
& $\mathbf{83.94 \pm 5.27}$
\\
& 70 & Clean
& $72.31 \pm 0.16$ 
& $68.91 \pm 0.09$ 
& $58.88 \pm 2.33$ 
& $70.45 \pm 0.05$ 
& $79.84 \pm 1.07$
& $\mathbf{87.79 \pm 2.45}$
\\\midrule
\multirow{15}{*}{SQV}
& 10 & -5 dB 
& ---
& ---
& $29.32 \pm 0.23$ 
& --- 
& $39.74 \pm 0.76$
& $\mathbf{41.54 \pm 0.75}$
\\
& 20 & -5 dB 
& $16.28 \pm 1.16$ 
& $27.82 \pm 0.19$ 
& $28.28 \pm 0.42$ 
& $23.66 \pm 0.35$ 
& $40.05 \pm 0.66$
& $\mathbf{52.80 \pm 1.56}$
\\
& 30 & -5 dB 
& $15.22 \pm 0.09$ 
& $27.73 \pm 0.14$ 
& $28.55 \pm 0.18$ 
& $23.86 \pm 0.93$ 
& $39.98 \pm 0.65$
& $\mathbf{54.87 \pm 1.69}$
\\
& 40 & -5 dB 
& $15.35 \pm 1.00$ 
& $26.54 \pm 0.27$ 
& $28.57 \pm 0.12$ 
& $24.78 \pm 0.29$ 
& $39.12 \pm 0.87$
& $\mathbf{58.90 \pm 1.68}$
\\
& 50 & -5 dB 
& $15.71 \pm 0.56$ 
& $24.85 \pm 0.09$ 
& $28.52 \pm 0.32$ 
& $24.82 \pm 0.27$ 
& $39.95 \pm 0.20$
& $\mathbf{60.64 \pm 1.04}$
\\
& 60 & -5 dB 
& $18.04 \pm 1.35$ 
& $25.48 \pm 0.22$ 
& $28.59 \pm 0.18$ 
& $24.52 \pm 0.57$ 
& $41.54 \pm 1.41$
& $\mathbf{61.21 \pm 0.78}$
\\
& 70 & -5 dB 
& $16.41 \pm 0.61$ 
& $27.47 \pm 0.09$ 
& $28.63 \pm 0.16$ 
& $24.40 \pm 0.44$ 
& $41.43 \pm 0.43$
& $\mathbf{61.98 \pm 1.46}$
\\
\\
& 10 & Clean
& --- 
& ---
& $73.85 \pm 0.23$ 
& --- 
& $79.18 \pm 0.90$
& $\mathbf{79.67 \pm 0.89}$
\\
& 20 & Clean
& $65.70 \pm 1.99$ 
& $72.31 \pm 0.13$ 
& $74.23 \pm 0.16$ 
& $72.22 \pm 0.09$ 
& $79.05 \pm 1.14$
& $\mathbf{85.51 \pm 1.03}$
\\
& 30 & Clean
& $65.31 \pm 1.22$ 
& $71.63 \pm 0.14$
& $73.96 \pm 0.12$ 
& $72.51 \pm 0.05$ 
& $79.38 \pm 0.89$
& $\mathbf{87.91 \pm 0.91}$
\\
& 40 & Clean
& $62.62 \pm 0.99$ 
& $70.82 \pm 0.63$ 
& $73.96 \pm 0.09$ 
& $72.34 \pm 0.09$ 
& $79.49 \pm 1.00$
& $\mathbf{90.62 \pm 0.26}$
\\
& 50 & Clean
& $69.25 \pm 2.59$ 
& $69.49 \pm 0.09$ 
& $74.16 \pm 0.03$ 
& $72.42 \pm 0.09$ 
& $80.16 \pm 1.41$
& $\mathbf{91.30 \pm 0.57}$
\\
& 60 & Clean
& $68.24 \pm 3.36$ 
& $69.32 \pm 0.07$ 
& $73.94 \pm 0.09$ 
& $72.23 \pm 0.10$ 
& $80.20 \pm 0.92$
& $\mathbf{92.31 \pm 0.47}$
\\
& 70 & Clean
& $71.23 \pm 3.20$ 
& $71.54 \pm 0.43$ 
& $74.12 \pm 0.21$ 
& $72.45 \pm 0.18$ 
& $80.64 \pm 1.08$
& $\mathbf{94.03 \pm 0.27}$
\\\bottomrule 
\end{tabular}
\label{table:tab_perf_results}
\end{adjustbox}
\end{table*}
Finally, we visualize the prediction capabilities of the different algorithms using t-Distributed Stochastic Neighbor Embedding (t-SNE) \cite{van2008visualizing} as shown in Figs. \ref{fig:t-sne_KAIST} and \ref{fig:t-sne_SQV}. The Dual-Domain Fusion is the only method than can somewhat clearly separate the Normal and Outer classes on KAIST. However, FixMatch, MixMatch, and Adversarial training, and Self-training seem to be better at separating \emph{Normal} and \emph{Inner} classes. Thus, further performance gains could be achieved with model ensembling. The major difference on SQV is that our Dual-Domain Fusion is able to separate the \emph{Normal} and \emph{Outer 1} classes from the other classes unlike the other methods.

\section{Conclusions} \label{sec:conclusions}
We proposed Dual-Domain Fusion, a new model-agnostic semi-supervised learning framework applicable to any one-dimensional time-series signal. The framework performs dual-domain training by combining the one-dimensional time-domain signals with their two-dimensional time–frequency representations and fusing them to maximize learning performance. Our deployment-aware architecture introduces no additional inference-time cost as all new samples are predicted solely from the time-domain branch. Experimental results on two public fault diagnosis datasets show substantial accuracy improvements of 8–46\% over FixMatch, MixMatch, Mean Teacher, Adversarial Training, and Self-training.

The classification models utilized in the reported experiments are simple, albeit effective, and more sophisticated techniques, such as large language models, could improve the performance even further along with model ensembling. Our work could also benefit from employing transfer learning techniques and optimizing the time–frequency representations. The experiments in this work focused on the bearing fault diagnosis application, but it would be interesting to see how the proposed method would perform on a wider set of other semi-supervised time-series classification, regression or domain adaptation problems in the future.
\bibliography{Sections/references}

\begin{thebibliography}{10}
\providecommand{\url}[1]{#1}
\csname url@samestyle\endcsname
\providecommand{\newblock}{\relax}
\providecommand{\bibinfo}[2]{#2}
\providecommand{\BIBentrySTDinterwordspacing}{\spaceskip=0pt\relax}
\providecommand{\BIBentryALTinterwordstretchfactor}{4}
\providecommand{\BIBentryALTinterwordspacing}{\spaceskip=\fontdimen2\font plus
\BIBentryALTinterwordstretchfactor\fontdimen3\font minus \fontdimen4\font\relax}
\providecommand{\BIBforeignlanguage}[2]{{%
\expandafter\ifx\csname l@#1\endcsname\relax
\typeout{** WARNING: IEEEtran.bst: No hyphenation pattern has been}%
\typeout{** loaded for the language `#1'. Using the pattern for}%
\typeout{** the default language instead.}%
\else
\language=\csname l@#1\endcsname
\fi
#2}}
\providecommand{\BIBdecl}{\relax}
\BIBdecl

\bibitem{chen2023deep}
X.~Chen, R.~Yang, Y.~Xue, M.~Huang, R.~Ferrero, and Z.~Wang, ``Deep transfer learning for bearing fault diagnosis: A systematic review since 2016,'' \emph{IEEE Transactions on Instrumentation and Measurement}, 2023.

\bibitem{qi2020small}
G.-J. Qi and J.~Luo, ``Small data challenges in big data era: A survey of recent progress on unsupervised and semi-supervised methods,'' \emph{IEEE Transactions on Pattern Analysis and Machine Intelligence}, vol.~44, no.~4, pp. 2168--2187, 2020.

\bibitem{gao2024multi}
T.~Gao, J.~Yang, and Q.~Tang, ``A multi-source domain information fusion network for rotating machinery fault diagnosis under variable operating conditions,'' \emph{Information Fusion}, p. 102278, 2024.

\bibitem{wang2024self}
L.~Wang, Y.~Gao, X.~Li, and L.~Gao, ``Self-supervised-enabled open-set cross-domain fault diagnosis method for rotating machinery,'' \emph{IEEE Transactions on Industrial Informatics}, 2024.

\bibitem{tang2024fault}
Z.~Tang, Z.~Su, S.~Wang, M.~Luo, H.~Luo, and L.~Bo, ``Fault diagnosis of rotating machinery toward unseen working condition: A regularized domain adaptive weight optimization,'' \emph{IEEE Transactions on Industrial Informatics}, 2024.

\bibitem{jalonen2023bearing_fault}
T.~Jalonen, M.~Al-Sa'd, S.~Kiranyaz, and M.~Gabbouj, ``Real-time vibration-based bearing fault diagnosis under time-varying speed conditions,'' in \emph{2024 IEEE International Conference on Industrial Technology (ICIT)}.\hskip 1em plus 0.5em minus 0.4em\relax IEEE, 2024, pp. 1--7.

\bibitem{wang2019understanding}
H.~Wang, Z.~Liu, D.~Peng, and Y.~Qin, ``Understanding and learning discriminant features based on multiattention 1dcnn for wheelset bearing fault diagnosis,'' \emph{IEEE Transactions on Industrial Informatics}, vol.~16, no.~9, pp. 5735--5745, 2019.

\bibitem{wang2025semi}
G.~Wang, C.~Pu, D.~Fu, Y.~Zhang, J.~Yu, and Y.~Hou, ``Semi-supervised federated learning fault diagnosis method driven by teacher-student model consistency,'' \emph{Signal, Image and Video Processing}, vol.~19, no.~5, p. 385, 2025.

\bibitem{luo2024fft}
X.~Luo, H.~Wang, T.~Han, and Y.~Zhang, ``Fft-trans: Enhancing robustness in mechanical fault diagnosis with fourier transform-based transformer under noisy conditions,'' \emph{IEEE Transactions on Instrumentation and Measurement}, 2024.

\bibitem{shi2021novel}
Z.~Shi, J.~Chen, Y.~Zi, and Z.~Zhou, ``A novel multitask adversarial network via redundant lifting for multicomponent intelligent fault detection under sharp speed variation,'' \emph{IEEE Transactions on Instrumentation and Measurement}, vol.~70, pp. 1--10, 2021.

\bibitem{alsad2024quadratic}
M.~Al-Sa'd, T.~Jalonen, S.~Kiranyaz, and M.~Gabbouj, ``Quadratic time-frequency analysis of vibration signals for diagnosing bearing faults,'' \emph{arXiv preprint arXiv:2401.01172}, 2024.

\bibitem{qu2023adaptive}
Y.~Qu, X.~Wang, X.~Zhang, and S.~Huang, ``An adaptive method for multifault diagnosis of induction motor under sharp changing speed and load condition,'' \emph{IEEE Transactions on Industrial Informatics}, 2023.

\bibitem{deng2021double}
Y.~Deng, D.~Huang, S.~Du, G.~Li, C.~Zhao, and J.~Lv, ``A double-layer attention based adversarial network for partial transfer learning in machinery fault diagnosis,'' \emph{Computers in Industry}, vol. 127, p. 103399, 2021.

\bibitem{verstraete2020deep}
D.~B. Verstraete, E.~L. Droguett, V.~Meruane, M.~Modarres, and A.~Ferrada, ``Deep semi-supervised generative adversarial fault diagnostics of rolling element bearings,'' \emph{Structural Health Monitoring}, vol.~19, no.~2, pp. 390--411, 2020.

\bibitem{9456035}
M.~Al-Sa’d, B.~Boashash, and M.~Gabbouj, ``{Design of an Optimal Piece-Wise Spline Wigner-Ville Distribution for TFD Performance Evaluation and Comparison},'' \emph{IEEE Transactions on Signal Processing}, vol.~69, pp. 3963--3976, 2021.

\bibitem{TFBook}
B.~Boashash, Ed., \emph{{Time-Frequency Signal Analysis and Processing}}, 2nd~ed.\hskip 1em plus 0.5em minus 0.4em\relax Oxford: Academic Press, 2016.

\bibitem{zheng2021self}
S.~Zheng and J.~Zhao, ``A self-adaptive temporal-spatial self-training algorithm for semisupervised fault diagnosis of industrial processes,'' \emph{IEEE transactions on industrial informatics}, vol.~18, no.~10, pp. 6700--6711, 2021.

\bibitem{sohn2020fixmatch}
K.~Sohn, D.~Berthelot, N.~Carlini, Z.~Zhang, H.~Zhang, C.~A. Raffel, E.~D. Cubuk, A.~Kurakin, and C.-L. Li, ``Fixmatch: Simplifying semi-supervised learning with consistency and confidence,'' \emph{Advances in neural information processing systems}, vol.~33, pp. 596--608, 2020.

\bibitem{berthelot2019mixmatch}
D.~Berthelot, N.~Carlini, I.~Goodfellow, N.~Papernot, A.~Oliver, and C.~A. Raffel, ``Mixmatch: A holistic approach to semi-supervised learning,'' \emph{Advances in neural information processing systems}, vol.~32, 2019.

\bibitem{tarvainen2017mean}
A.~Tarvainen and H.~Valpola, ``Mean teachers are better role models: Weight-averaged consistency targets improve semi-supervised deep learning results,'' \emph{Advances in neural information processing systems}, vol.~30, 2017.

\bibitem{miyato2018virtual}
T.~Miyato, S.-i. Maeda, M.~Koyama, and S.~Ishii, ``Virtual adversarial training: a regularization method for supervised and semi-supervised learning,'' \emph{IEEE transactions on pattern analysis and machine intelligence}, vol.~41, no.~8, pp. 1979--1993, 2018.

\bibitem{yan2022cotraining}
X.~Yan, X.~Xia, L.~Wang, and Z.~Zhang, ``A cotraining-based semisupervised approach for remaining-useful-life prediction of bearings,'' \emph{Sensors}, vol.~22, no.~20, p. 7766, 2022.

\bibitem{liu2023temporal}
Z.~Liu, Q.~Ma, P.~Ma, and L.~Wang, ``Temporal-frequency co-training for time series semi-supervised learning,'' in \emph{Proceedings of the AAAI conference on artificial intelligence}, vol.~37, no.~7, 2023, pp. 8923--8931.

\bibitem{liang2025novel}
Q.~Liang, X.~Liang, M.~Zhang, C.~Liu, and T.~Zhang, ``A novel comprehensive semi-supervised learning method for fault diagnosis under extremely low label rate,'' \emph{IEEE Transactions on Instrumentation and Measurement}, 2025.

\bibitem{razavi2018information}
R.~Razavi-Far, E.~Hallaji, M.~Farajzadeh-Zanjani, M.~Saif, S.~H. Kia, H.~Henao, and G.-A. Capolino, ``Information fusion and semi-supervised deep learning scheme for diagnosing gear faults in induction machine systems,'' \emph{IEEE Transactions on Industrial Electronics}, vol.~66, no.~8, pp. 6331--6342, 2018.

\bibitem{wei2023time}
C.~Wei, Z.~Wang, J.~Yuan, C.~Li, and S.~Chen, ``Time-frequency based multi-task learning for semi-supervised time series classification,'' \emph{Information Sciences}, vol. 619, pp. 762--780, 2023.

\bibitem{guo2020online}
Q.~Guo, X.~Wang, Y.~Wu, Z.~Yu, D.~Liang, X.~Hu, and P.~Luo, ``Online knowledge distillation via collaborative learning,'' in \emph{Proceedings of the IEEE/CVF Conference on Computer Vision and Pattern Recognition}, 2020, pp. 11\,020--11\,029.

\bibitem{xu2022intelligent}
Z.~Xu, M.~Bashir, W.~Zhang, Y.~Yang, X.~Wang, and C.~Li, ``An intelligent fault diagnosis for machine maintenance using weighted soft-voting rule based multi-attention module with multi-scale information fusion,'' \emph{Information Fusion}, vol.~86, pp. 17--29, 2022.

\bibitem{yang2023decision}
A.~Yang, M.~Wu, C.~Lu, W.~Yu, J.~Hu, and Y.~Nakanishi, ``Decision fusion scheme based on mode decomposition and evidence theory for fault diagnosis of drilling process,'' \emph{IEEE Transactions on Industrial Informatics}, 2023.

\bibitem{fan2023conservative}
S.~Fan, F.~Zhu, Z.~Feng, Y.~Lv, M.~Song, and F.-Y. Wang, ``Conservative-progressive collaborative learning for semi-supervised semantic segmentation,'' \emph{IEEE Transactions on Image Processing}, 2023.

\bibitem{ye2024semi}
T.~Ye, X.~Yuan, X.~Yang, Y.~Song, Z.~Zhang, and F.~Zhou, ``A semi-supervised intelligent fault diagnosis method for bearings under low labeled rates,'' \emph{IEEE Transactions on Instrumentation and Measurement}, 2024.

\bibitem{KAIST_data}
W.~Jung, S.-H. Kim, S.-H. Yun, J.~Bae, and Y.-H. Park, ``\BIBforeignlanguage{eng}{Vibration, acoustic, temperature, and motor current dataset of rotating machine under varying operating conditions for fault diagnosis},'' \emph{\BIBforeignlanguage{eng}{Data in brief}}, vol.~48, pp. 109\,049--109\,049, 2023.

\bibitem{liu2022subspace}
S.~Liu, J.~Chen, S.~He, Z.~Shi, and Z.~Zhou, ``Subspace network with shared representation learning for intelligent fault diagnosis of machine under speed transient conditions with few samples,'' \emph{ISA transactions}, vol. 128, pp. 531--544, 2022.

\bibitem{van2008visualizing}
L.~van~der Maaten and G.~Hinton, ``{Visualizing data using t-SNE},'' \emph{Journal of Machine Learning Research}, vol.~9, no.~86, pp. 2579--2605, 2008.

\end{thebibliography}
\end{document}